\documentclass[conference]{IEEEtran}
\IEEEoverridecommandlockouts

\usepackage{cite}
\usepackage{svg}
\usepackage{amsfonts}
\usepackage{algorithmic}
\usepackage{graphicx}
\usepackage{textcomp}
\usepackage[table]{xcolor}
\usepackage{xcolor}
\usepackage{float}
\usepackage{amsmath}
\usepackage{algorithm} 
\usepackage{amssymb}
\usepackage{enumitem}
\usepackage{booktabs}
\usepackage{array}
\usepackage{tabularx}
\usepackage{orcidlink} 
\usepackage{hyperref}
\usepackage{xcolor}
\newcolumntype{V}{!{\vrule width 1pt}}
\def\BibTeX{{\rm B\kern-.05em{\sc i\kern-.025em b}\kern-.08em
    T\kern-.1667em\lower.7ex\hbox{E}\kern-.125emX}}
\begin{document}

\title{TS-Fault: Benchmarking Time Series Forecasters Against Structural Faults 

\author{
Yuyang Zhao$^{1}$, Lian Xu$^{2}$, Hao Miao$^{3}$, Chenxi Liu$^{4}$, and Hao Xue$^{1\boxtimes}$\\
\textit{$^{1}$Hong Kong University of Science and Technology (Guangzhou), Guangzhou, China}\\
\textit{$^{2}$The University of Western Australia, Perth, Australia}\\
\textit{$^{3}$Hong Kong Polytechnic University, Hong Kong, China}\\
\textit{$^{4}$CAIR, Hong Kong Institute of Science \& Innovation, Chinese Academy of Sciences}\\
\{yuyangzhao, haoxue\}@hkust-gz.edu.cn, lian.xu@uwa.edu.au, hao.miao@polyu.edu.hk, chenxi.liu@cair-cas.org.hk\\

}

\thanks{$^{\boxtimes}$Corresponding author.}
}

\maketitle
\begin{abstract}
Time series forecasting (TSF) underpins consequential decisions in energy, transportation, finance, and healthcare, yet TSF models are almost universally ranked by a single number (\textit{e.g.,} average error) on clean held-out data, under the implicit assumption that it predicts deployed reliability. However, real faults are not i.i.d.\ noise but structured events with temporal shape, broken cross-variable dependencies, regime change coupled with missingness, and causal propagation across a sensing pipeline. Treating TSF robustness as a data-quality problem, we present TS-Fault, a benchmark that evaluates forecasting models under explicit, parameterized fault scenarios with controllable semantic difficulty. TS-Fault organizes recurring failures into four modes along two orthogonal axes (observation- vs.\ mechanism-level; univariate vs.\ multivariate) and injects each fault into the most prediction-critical window via a unified importance score. This design enables robustness to be tested against the structures models actually rely on, rather than reduced to generic noise sensitivity. We evaluate 21 models across 6 datasets, 4 modes, and 5 difficulty levels under a paired clean/corrupt protocol. The results reveal three findings that contradict common leaderboard intuition: \emph{(i)} clean-data accuracy anti-correlates with robustness; \emph{(ii)} clean rankings are preserved under observation-level faults but reshuffled under mechanism-level faults; and \emph{(iii)} all catastrophic failures occur under mechanism-level faults, with foundation models achieving the highest clean-data accuracy yet exhibiting the greatest fragility. The code is publicly available at~\href{https://github.com/Ray-zyy/TS-Fault}{https://github.com/Ray-zyy/TS-Fault}.
\end{abstract}


\section{Introduction}
The proliferation of edge computing and mobile sensing generates a massive volume of time series data, which is being collected and stored in time series database systems~\cite{jensen2018modelardb, yao2024camel}, motivating various real-world applications~\cite{vlahogianni2014short, zhou2022fedformer}, \textit{e.g.,} time series forecasting~\cite{qiu2024tfb}. 
The forecasting is rarely an end in itself. In energy dispatch~\cite{hong2016probabilistic,nowotarski2018recent}, clinical monitoring~\cite{shickel2017deep,morid2023time}, financial risk control~\cite{sezer2020financial}, and traffic management~\cite{vlahogianni2014short}, a forecast is an input to a consequential downstream action, whose reliability rests on an assumption that current evaluation rarely tests: that the error a model exhibits during evaluation is representative of the error it will produce after deployment. Mainstream TSF evaluation has crystallized this assumption. For over two decades, progress in long-term TSF has been measured by a single family of quantities, \textit{i.e.,} average MSE/MAE on clean, complete, evenly sampled held-out series, and successive benchmark generations (M3/M4~\cite{makridakis2000m3,makridakis2020m4}, Monash~\cite{godahewa2021monash}, GIFT-Eval~\cite{aksu2024gift}, TFB~\cite{qiu2024tfb}) have largely refined how this quantity is measured rather than questioning whether it reflects deployment risk. In short, existing leaderboards answer ``which model is most accurate on clean data?'' whereas deployment asks a different question: ``under what conditions, and how severely, will a model fail?''

These two questions are not equivalent, and the gap is large enough to invert model-selection decisions. In our study, the model with the second-lowest error on clean data collapses to the worst-performing model under structured faults, while several models with only middling performance on clean data emerge as the most robust. Selecting a forecaster by MSE on clean data, the prevailing practice, can therefore systematically favor the model that is most brittle in deployment, and the current evaluation paradigm is unable to reveal this in advance.

Why the gap persists in part because existing robustness studies model a fault as a value-level deviation from otherwise nominal observations: additive Gaussian noise, random masking~\cite{cheng2024robusttsf,du2023saits}, or $\epsilon$-bounded adversarial perturbations~\cite{goodfellow2014explaining,madry2017towards}. These deviations typically assume that corruptions are independent across time and across variables, and approximately homogeneous in distribution. As a result, they are unable to express the faults that dominate real deployments. Real-world faults often violate these assumptions. A frozen sensor does not emit white noise; it emits an event with onset, peak, and decay, possibly displaced in time by a buffering pipeline. A market shock rewrites the lead–lag and gain structure among assets rather than perturbing each one independently~\cite{forbes2002no}, even while each series, viewed alone, still looks plausible. When a grid enters emergency operation, its monitoring drops samples in state-dependent, block-structured runs concentrated around the critical transition, not at random~\cite{busby2021cascading}. And upstream sensor drift propagates along causal dependencies instead of staying local,
dragging downstream true states with it and inducing secondary observation failures. None of these mechanisms can be faithfully expressed by i.i.d.\ perturbations of a clean signal, no matter how the variance or masking rate is changed. Evaluating TSF reliability should therefore be viewed as a data-quality problem involving structured dirty data, rather than as a question of sensitivity to additive noise.

We argue that closing this gap requires changing the object of evaluation, not merely adding harder samples. 
We present TS-Fault, a benchmark that evaluates time series forecasters under explicit, parameterized fault scenarios rather than clean test inputs alone. Each scenario specifies what type of fault occurs, where it occurs, how severe it is, and which temporal or cross-variable structures it disrupts. This makes the fault itself a first-class object of evaluation, enabling model degradation to be interpreted in terms of named failure mechanisms rather than reported only as an aggregate error increase.
We organize recurring real-world failures into four fault modes along two orthogonal axes: whether a fault corrupts observations (a transient event) or alters the data-generating mechanism (a persistent regime), and whether it acts on a single series or on the cross-variable structure. Crucially, every fault is injected not at a random position but into the most prediction-critical window, selected by a unified importance score, so that the benchmark stresses the regions a model actually relies on rather than degenerating into a random perturbation. 
Our main contributions can be summarized as follows:

\begin{itemize}[leftmargin=*]
\item \emph{A fault-operator framework} (Section~\ref{sec:framework}) that reformulates TSF robustness evaluation around explicit fault scenarios, semantic fault parameters, controllable difficulty levels, worst-case and average risk measures, and a unified strategy for identifying prediction-critical windows.
\item \emph{Four parameterized fault modes} (Section~\ref{sec:modes}) spanning the $2\times2$ taxonomy, each with a concrete, causally grounded construction and an explicit difficulty decomposition: Time-Warped Shock (Mode~I), Dependency-Fracture Shock (Mode~II), Regime-Transition Missingness (Mode~III), and Cascading Sensor-to-System Failure (Mode~IV).
\item \emph{A reproducible benchmark and large-scale empirical study} covering 21 models, 6 datasets, 4 modes, and 5 difficulty levels under a paired clean/corrupt protocol. The results reveal three findings that challenges standard leaderboard intuition: (\textbf{i}) clean-data performance anti-correlates with robustness; (\textbf{ii}) a sharp stratification of failure impact across modes; (\textbf{iii}) a concentration of all catastrophic failures in mechanism-level modes, with pretrained foundation models being the most accurate yet the most fragile.
\item \emph{A diagnostic view of forecasting robustness} that attributes model degradation to specific fault mechanisms and difficulty levels. Rather than treating robustness as a pass/fail noise test, TS-Fault enables ablation-style analysis of when, how, and why forecasters fail under structured faults.
\end{itemize}

\section{Related Work}
\label{sec:background}

\subsection{TSF Benchmarks and Evaluation Paradigms}
Time series forecasting (TSF) has been evaluated for decades through benchmark protocols that rank models by aggregate forecasting error on held-out data, as shown in Table~\ref{tab:generations}. The first generation (M3, M4~\cite{makridakis2000m3,makridakis2020m4}) established accuracy comparison across heterogeneous domains; the second (Monash~\cite{godahewa2021monash} and the LTSF suite~\cite{zhou2021informer}) standardized cross-domain and long-horizon multivariate protocols; the third (GIFT-Eval~\cite{aksu2024gift}, TFB~\cite{qiu2024tfb}, BasicTS+~\cite{shao2024exploring}) tightened held-out discipline and fairness of comparison; and the fourth (fev-bench~\cite{shchur2025fev}, TSFM-Bench~\cite{li2025tsfm}, ProbTS~\cite{zhang2024probts}), responding to pretrained forecasters, acknowledged that pretraining leakage can no longer be reliably excluded. 


\begin{table}[t]
\centering
\caption{Generational evolution of TSF benchmarks.}
\label{tab:generations}
\footnotesize
\begin{tabular}{lll}
\toprule
\textbf{Gen.} & \textbf{Representative} & \textbf{Treatment of faults} \\
\midrule
G1 (2000--18) & M3, M4            & Implicit, via domain diversity \\
G2 (2018--22) & Monash, LTSF        & Explicit domain-shift test \\
G3 (2023--24) & GIFT-Eval, TFB      & Controlled by held-out discipline \\
G4 (2024--25) & fev-bench, ProbTS   & Acknowledged but unmitigated \\
\textbf{Ours} & \textbf{TS-Fault}   & \textbf{Primary design variable} \\
\bottomrule
\end{tabular}
\end{table}

Despite these improvements, evaluation remains centered on average MSE or MAE on clean held-out data. This scalar ranking reveals which model is accurate, but not when, how severely, or why it fails. In contrast, TS-Fault evaluates models under explicit fault conditions, reporting degradation by failure mode and difficulty level and attributing each degradation to a named, parameterized mechanism.



\subsection{Time Series Foundation Models and Their Evaluation}
\label{sec:tsfm}
The evaluation problem has become more urgent with the rise of the time series foundation
model (TSFM). Trained on large, heterogeneous corpora~\cite{kaplan2020scaling,hoffmann2022training} and applied zero-shot, models such as Chronos~\cite{ansari2024chronos}, TimesFM~\cite{das2023decoder} and Moirai~\cite{woo2024unified} attain strong clean-data accuracy without task-specific training~\cite{liang2024foundation}. However, foundation models also complicate evaluation. Because their pretraining corpora are large, only partially auditable, and often overlap the public repositories that later supply benchmark test sets, strong held-out performance becomes hard to distinguish from memorization or implicit exposure to benchmark data~\cite{sainz2023nlp,golchin2023time,xu2024benchmark}. Existing benchmark responses mainly tighten held-out protocols, require disclosure of pretraining corpora, or introduce more careful dataset filtering. These measures protect the interpretation of a clean score but leave a prior question unanswered: how such models behave when their input is structurally faulted. 
TS-Fault directly addresses this question by providing a complementary evaluation axis by synthesizing faulted instances at evaluation time from controlled scenario parameters, so that even a memorized clean sequence does not directly provide the structured failure pattern used for testing (Sec.~\ref{sec:rq4}).

\begin{figure*}[t]
\centering
\includegraphics[width=.95\textwidth]{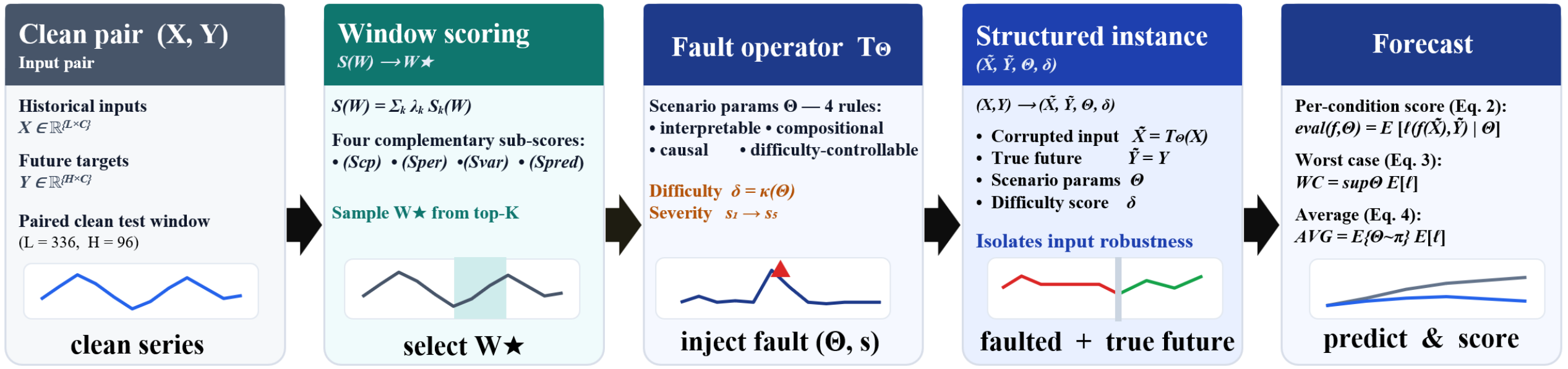}
\caption{The TS-Fault pipeline.}
\label{fig:workflow}
\end{figure*}

\subsection{Robustness Evaluation in Time Series}
\label{sec:imagenetc} 
Existing robustness research in time series can be organized along two complementary axes: whether a perturbation is sampled from a distribution or constructed by an explicit operator, and whether it is unstructured (i.i.d.\ across time and variables) or structured (carrying temporal shape, cross-variable coupling, or causal propagation). Table~\ref{tab:quadrants} summarizes this landscape. Noise and masking assume i.i.d.\ timesteps and cannot express cross-temporal failure structure~\cite{cheng2024robusttsf,du2023saits,cao2018brits}; adversarial perturbations are mathematical objects within $\epsilon$-balls that lack semantic grounding~\cite{goodfellow2014explaining,madry2017towards}; and distribution-shift benchmarks, though closest to our concern, remain observational, they measure degradation across time gaps without enabling counterfactual control over how the same shift would behave if made more severe~\cite{yao2022wild,gagnon2022woods,koh2021wilds}. The remaining quadrant (constructed and structured), is largely unexplored in TSF robustness evaluation. This is most relevant to deployment faults, where failures often have temporal morphology, cross-variable dependence, state dependence, and propagation effects. TS-Fault fills this gap by introducing explicit, scenario-grounded fault operators with controllable difficulty. This follows the general philosophy of corruption-robustness benchmarks such as ImageNet-C~\cite{hendrycks2019benchmarking}, but adapts it to the distinctive structure of time series, where robustness depends on how faults unfold across time and variables, not just on corruption intensity.

\begin{table}[t]
\centering
\caption{Four quadrants of TSF robustness research.}
\label{tab:quadrants}
\footnotesize
\renewcommand{\arraystretch}{1.08}
\setlength{\tabcolsep}{2.5pt}
\begin{tabularx}{\columnwidth}{@{}l
  >{\raggedright\arraybackslash}X
  >{\raggedright\arraybackslash}X@{}}
\toprule
 & \textbf{Sampled} & \textbf{Constructed} \\
\midrule
\textbf{Unstructured}
& Noise \& missingness (Gaussian, masking)~\cite{cheng2024robusttsf,du2023saits,cao2018brits}
& Adversarial attacks (FGSM/PGD in $\epsilon$-balls)~\cite{goodfellow2014explaining,madry2017towards} \\
\textbf{Structured}
& Distribution shift (Wild-Time~\cite{yao2022wild}, drift)~\cite{gagnon2022woods,koh2021wilds}
& \textbf{Scenario-grounded (TS-Fault, this work)} \\
\bottomrule
\end{tabularx}

\end{table}



\section{The Fault-Operator Framework}
\label{sec:framework}


This section develops an alternative to clean-accuracy evaluation in four steps. Sec.~\ref{sec:instances} defines fault operators and the structured test instance they produce, together with two risk measures over them. Sec.~\ref{sec:difficulty} constrains what may serve as a scenario parameter and grades it with a difficulty map. Sec.~\ref{sec:window} places each fault in the window a model actually relies on. Sec.~\ref{sec:generation} ties these components into a reproducible generator. Figure~\ref{fig:workflow} summarizes the resulting pipeline. The main idea of this pipeline is to introduce a transformation that injects a specific fault mechanism into clean data based on explicit and interpretable parameters.

\subsection{Structured Instances and Risk Measures}\label{sec:instances}
\noindent\textbf{Definition 1 (Fault operator).}\;
A \emph{fault operator} is a map $\mathcal{T}_\Theta:\mathbb{R}^{L\times C}\!\to\!\mathbb{R}^{L\times C}$ parameterized by a scenario-parameter vector $\Theta$. Applied to a clean context $X$, it produces a faulted context $\tilde{X}=\mathcal{T}_\Theta(X)$, where $\Theta$ encodes interpretable quantities such as fault onset, duration, affected channels, magnitude, and propagation range.

\noindent\textbf{Definition 2 (Fault family).}\;
A \emph{fault family} is a set of operators sharing one failure mechanism, $\mathcal{F}=\{\,\mathcal{T}_\Theta : \Theta\in\Phi\,\}$, where $\Phi$ is the parameter space of that mechanism. Distinct $\Theta\in\Phi$ instantiate the same mechanism under different conditions and difficulties.

We replace the clean pair $(X,Y)$ with a structured instance
\begin{equation}
(X,Y)\;\xrightarrow{\;\mathcal{T}_\Theta\;}\;\big(\tilde{X},\,\tilde{Y},\,\Theta,\,\delta\big),\qquad \tilde{X}=\mathcal{T}_\Theta(X),
\label{eq:instance}
\end{equation}
where the fault operator acts only on the input context, so the target is the clean, unperturbed future $\tilde{Y}=Y$ for all four modes.
This design isolates\emph{input} robustness. 
A model is always scored on recovering the true future from a corrupted history, never on predicting the corruption itself.
The scalar $\delta=\kappa(\Theta)$ grades how difficult the instance is (Sec.~\ref{sec:difficulty}).
Evaluation then becomes a function of $\Theta$:
\begin{equation}
\mathrm{eval}(f,\Theta)\;=\;\mathbb{E}\big[\,\ell\big(f(\tilde{X}),\tilde{Y}\big)
\,\big|\,\Theta\,\big].
\label{eq:cond}
\end{equation}
Hence, the performance is reported per fault condition, rather than as a single aggregate over an unspecified mixture of conditions.


\noindent\textbf{Risk measures.}\; Over a fault family $\mathcal{F}$, we define a worst-case and an average risk:
\begin{align}
\mathrm{WC}(f,\mathcal{F}) &= \sup_{\Theta\in\Phi}\;
\mathbb{E}_{(X,Y)\sim P}\big[\ell\big(f(\mathcal{T}_\Theta(X)),Y\big)\big],
\label{eq:wc}\\[2pt]
\mathrm{AVG}(f,\mathcal{F}) &= \mathbb{E}_{\Theta\sim\pi_\Phi}\;
\mathbb{E}_{(X,Y)\sim P}\big[\ell\big(f(\mathcal{T}_\Theta(X)),Y\big)\big].
\label{eq:avg}
\end{align}

$\mathrm{WC}$ bounds how bad it can get under the most adverse parameter. $\mathrm{AVG}$, taken under a prior $\pi_\Phi$ over the parameter space, is less sensitive to the exact parameterization and serves as a scalar summary for cross-model comparison. Reporting both separates ``how bad can it get'' from ``how bad is it on average.'' In this paper, all tabulated summaries report $\mathrm{AVG}$, whereas $\mathrm{WC}$ is reflected by the per-dataset maxima of Sec.~\ref{sec:rq6}.

\subsection{Scenario Parameters under a Difficulty Map}
\label{sec:difficulty}
Not every parameterized perturbation qualifies as a scenario parameter. We then require $\Theta$ to satisfy four properties. \emph{(i) Semantic interpretability:} every component corresponds to a quantity a domain expert recognizes (shock magnitude, event duration, coupling gain, sensor failure rate), rather than a coordinate in an abstract space such as a gradient step size. \emph{(ii) Compositionality:} components must combine to express compound conditions (\textit{e.g.}, a demand surge co-occurring with sensor dropout) without a fresh parameter vector per combination, enabling systematic coverage of the joint failure space. \emph{(iii) Causal grounding:} $\Theta$ must encode a mechanism~\cite{scholkopf2021toward,peters2017elements} rather than a statistical summary (\textit{i.e.}, the trigger, propagation path, and decay dynamics of a shift). \emph{(iv) Difficulty controllability:} the difficulty $\delta$ varies monotonically and predictably with $\Theta$, so a graded test can attribute a performance difference to a specific parameter.

\noindent\textbf{Difficulty score.}\; The difficulty $\delta=\kappa(\Theta)$ is a domain-interpretable mapping from scenario parameters to a scalar. Each mode decomposes $\delta$ into interpretable terms (event magnitude, structural distortion, information loss, coupling strength, see Sec.~\ref{sec:difficulty-detail}), so an instance reports not only that it is hard but why. 
Sweeping $\Theta$ along $\kappa$ yields graded difficulty levels, and with them robustness curves and threshold analyses instead of single-point comparisons.

\subsection{Localizing Faults: a Window-Importance Score}
\label{sec:window}
\begin{figure}[t]
\centering
\includegraphics[width=\columnwidth]{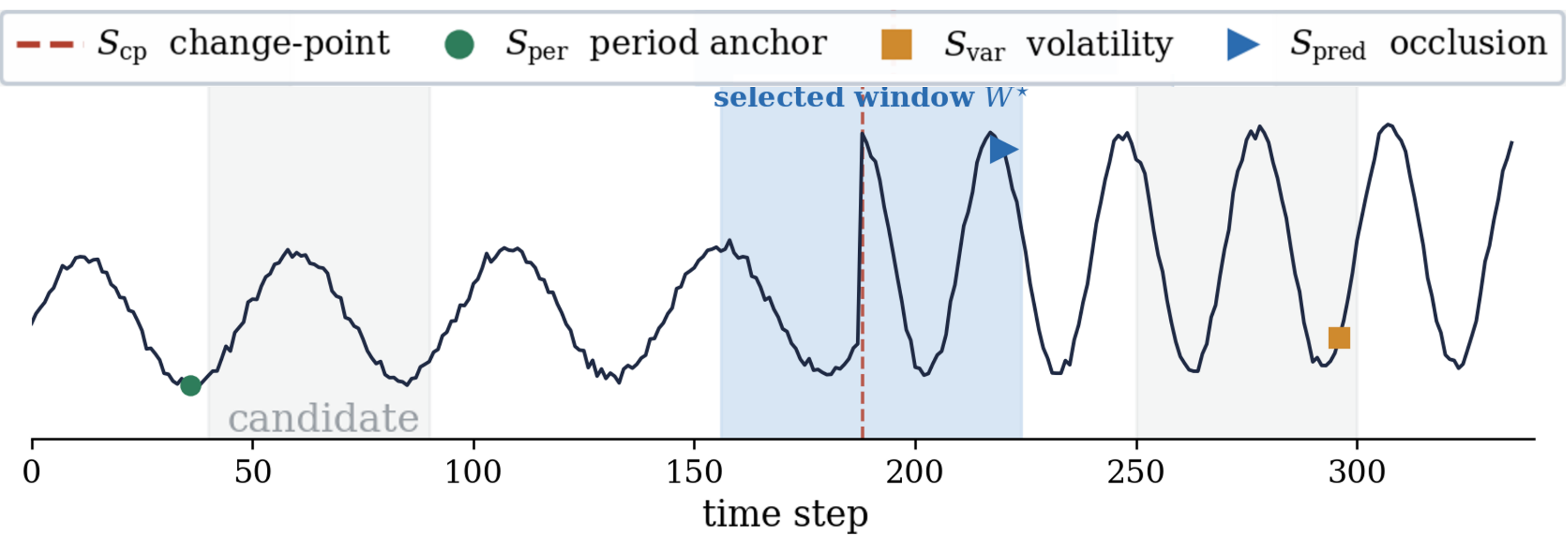}
\caption{Illustrative window-importance scoring. Among candidate windows (grey), $S(W)$ selects $W^\star$ (blue) because it is simultaneously close to a change point ($S_{\mathrm{cp}}$), near identifiable period anchors ($S_{\mathrm{per}}$), internally volatile ($S_{\mathrm{var}}$), and highly influential on the forecast under occlusion ($S_{\mathrm{pred}}$). Faults are injected here, not at random positions.}
\label{fig:windowimp}
\end{figure}

Injecting a fault at a random position would let the benchmark degenerate into a noise test. Hence, we further introduce a mechanism that selects the most prediction-critical window for all four modes. For a candidate window $W=[s,e]\subseteq[1,L]$ we compute:
\begin{equation}
\begin{aligned}
S(W)={}&\lambda_1 S_{\mathrm{cp}}(W)+\lambda_2 S_{\mathrm{per}}(W)\\
&+\lambda_3 S_{\mathrm{var}}(W)+\lambda_4 S_{\mathrm{pred}}(W),
\end{aligned}
\label{eq:window}
\end{equation}
and sample $W^\star$ from the top-$K$ candidates. The four sub-scores capture complementary reasons a model may rely on $W$, so that their sum favors windows that are critical for more than one reason (Fig.~\ref{fig:windowimp}).

A forecaster makes predictions based on the most recent behavior of the time series. When that behavior shifts, the window covering the shift is what tells the model which pattern currently holds, so a fault placed there would be especially challenging.
We use the change-point score $S_{\mathrm{cp}}$ to find such windows.
Thus, we split $W$ at its center $c$ into halves $W_L=[s,c]$ and $W_R=[c+1,e]$, fit a light local predictor on each half, and take the bidirectional cross-prediction error
\begin{equation}
\begin{aligned}
S_{\mathrm{cp}}(W)
&= \frac{1}{|W_R|}\sum_{t\in W_R}
\big\|x_t-f_L(x_{<t})\big\|_2^2 \\
&\quad + \frac{1}{|W_L|}\sum_{t\in W_L}
\big\|x_t-f_R(x_{>t})\big\|_2^2 .
\end{aligned}
\label{eq:scp}
\end{equation}
If $W$ does not cross a change, both halves follow the same local law, and each predictor explains the other half. The score will be high when $W$ covers a switch in trend, variance, or period~\cite{aminikhanghahi2017survey,keogh2005hot}.

Most forecasters rely heavily on periodicity, and a cycle is easiest to pin down at its peaks and valleys. The period score $S_{\mathrm{per}}$ is therefore introduced to target windows near these points.
We estimate the dominant period $P$ from the autocorrelation,
\begin{equation}
\begin{aligned}
\mathrm{ACF}(k)
&= \frac{\sum_{t}(x_t-\bar{x})(x_{t+k}-\bar{x})}
{\sum_{t}(x_t-\bar{x})^2}, \\
P
&= \arg\max_{2\le k\le L/2}\mathrm{ACF}(k).
\end{aligned}
\label{eq:acf}
\end{equation}
We then locate the peak/valley anchors of the nearest cycle and score the Gaussian-decayed closeness of the window center to the nearest anchor (bandwidth $\sigma_p$),
\begin{equation}
S_{\mathrm{per}}(W)=\max\!\Big\{e^{-d_{\mathrm{peak}}^2/\sigma_p^2},\,
e^{-d_{\mathrm{valley}}^2/\sigma_p^2}\Big\},
\label{eq:sper}
\end{equation}
so that windows aligned with a peak or a valley score highest.

A fault in a flat, uninformative stretch barely changes the forecast, so the volatility score $S_{\mathrm{var}}$ is also designed to measure how much information a window carries. With a wavelet decomposition of $W$ and per-level detail energy $E_j(W)=\sum_{t\in W}|d_{j,t}|^2$, we take the robustly normalized weighted sum (level weights $\omega_j$)
\begin{equation}
S_{\mathrm{var}}(W)=\frac{\sum_j \omega_j E_j(W)}{\mathrm{MAD}(W)+\varepsilon},
\label{eq:svar}
\end{equation}
favoring windows with rich internal dynamics, which tend to carry more predictive information.


The above three scores are computed from the data perspective, so they focus on approximating what a model actually uses.
The occlusion score $S_{\mathrm{pred}}$ measures this directly. With a reference predictor $f$, we compare the full-input forecast $\hat{y}$ to the forecast $\hat{y}^{(\backslash W)}$ obtained after occluding $W$,
\begin{equation}
S_{\mathrm{pred}}(W)=\sum_{h=1}^{H}\alpha_h\,
\big\|\hat{y}_{L+h}-\hat{y}^{(\backslash W)}_{L+h}\big\|_2^2,
\label{eq:spred}
\end{equation}
directly identifying the windows the model actually depends on. The weights $\lambda_i$ are per-mode hyperparameters (\textit{e.g.}, Mode~I raises $\lambda_3$ and $\lambda_4$ to favor internally dynamic windows with high predictive leverage, whereas Mode~III raises $\lambda_1$ to favor windows near a change point, detailed in Sec~\ref{sec:modes}). This could guarantee that the benchmark does not collapse into a random perturbation.

\subsection{Generating a TS-Fault Instance}
\label{sec:generation}
Algorithm~\ref{alg:gen} ties the components together. The same windowing mechanism (lines 2--3) is shared by all four modes, which is what makes their results directly comparable. Only the mode-specific structure selection (line~4) and the operator (line~5) differ. 

The generation cost is relatively cheap. With light local predictors, $S_{\mathrm{cp}}$ and $S_{\mathrm{pred}}$ are linear in the number of candidate windows, the remaining sub-scores are linear in window length, and all per-channel computations are independent, so generation scales linearly with the number of channels and parallelizes trivially across instances.
Because instances are produced by an explicit operator rather than sampled, the benchmark can also be regenerated at any difficulty by re-sweeping $\kappa$, and previously unexposed $\Theta$ combinations can be held out at release time (Sec.~\ref{sec:goodhart}).

\begin{algorithm}[t]
\caption{Generating a TS-Fault instance}
\label{alg:gen}
\begin{algorithmic}[1]
\REQUIRE clean window $X$, target $Y$; mode $i$; difficulty $s$
\STATE sample parameters $\Theta\sim\Phi_i$ at difficulty $s$ (i.e., subject to $\kappa_i(\Theta)=\delta_s$, the target difficulty of level $s$)
\STATE score candidates $S(W)\gets\sum_k \lambda^{(i)}_k S_k(W)$
\STATE sample the critical window $W^\star$ from the top-$K$ by $S(\cdot)$
\STATE select mode-specific structure on $W^\star$ (variable subset $S$; or root
       set $R$ and downstream $D$; or switch center $\tau$)
\STATE apply the operator, localized to $W^\star$: $\tilde{X}\gets \mathcal{T}_\Theta(X)$
\STATE $\tilde{Y}\gets Y$ \quad // target unchanged; only the input is faulted
\STATE $\delta\gets\kappa_i(\Theta)$
\STATE \textbf{return} structured instance $(\tilde{X},\tilde{Y},\Theta,\delta)$
\end{algorithmic}
\end{algorithm}

\begin{figure}[]
\centering
\includegraphics[width=\columnwidth]{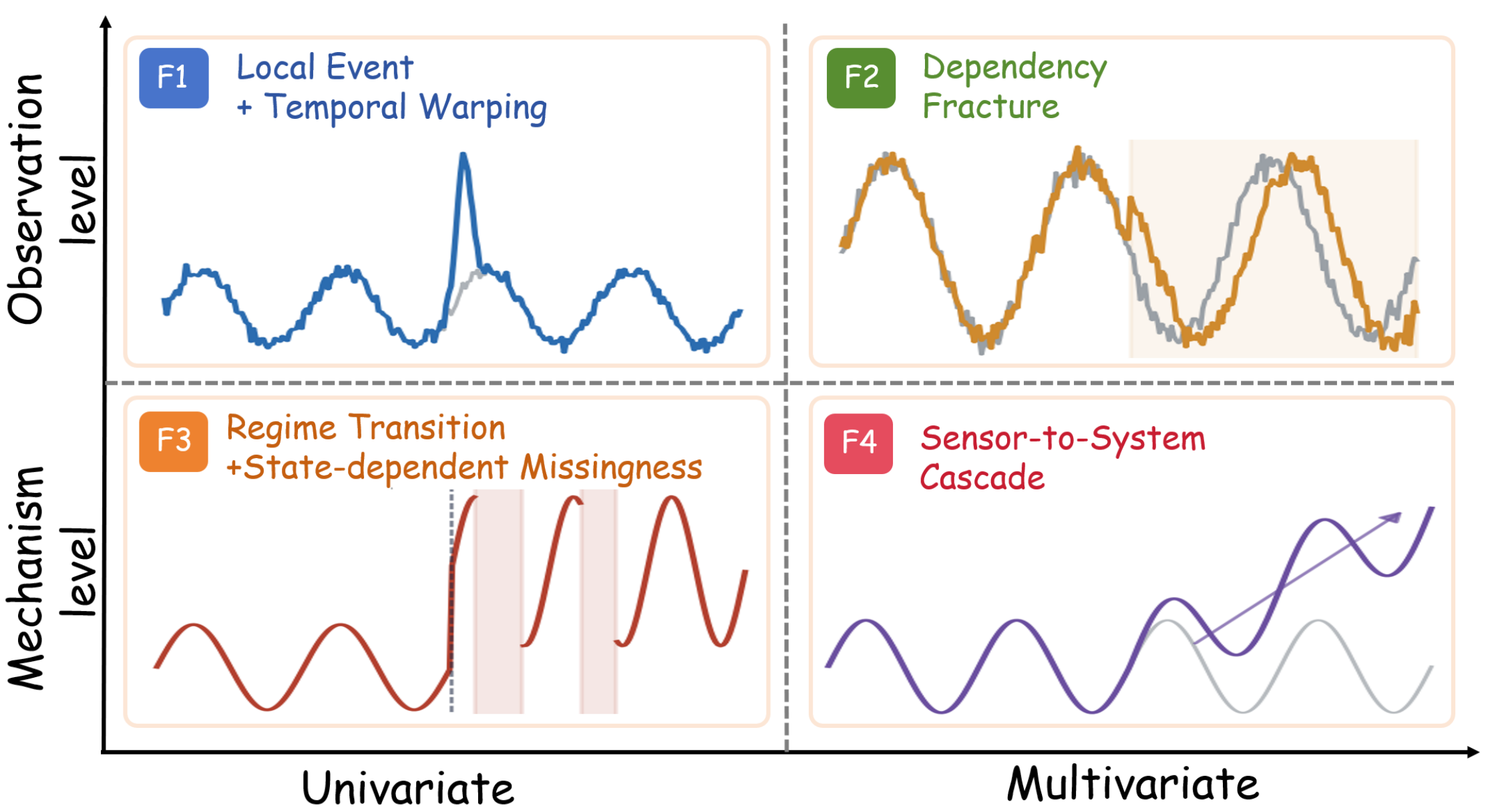}
\caption{The $2\times2$ fault taxonomy. Scope (observation- vs.\ mechanism-level)
$\times$ variate scope (uni- vs.\ multivariate) yields four modes. 
}
\label{fig:taxonomy}
\end{figure}

\begin{figure}[]
\centering
\includegraphics[width=\columnwidth]{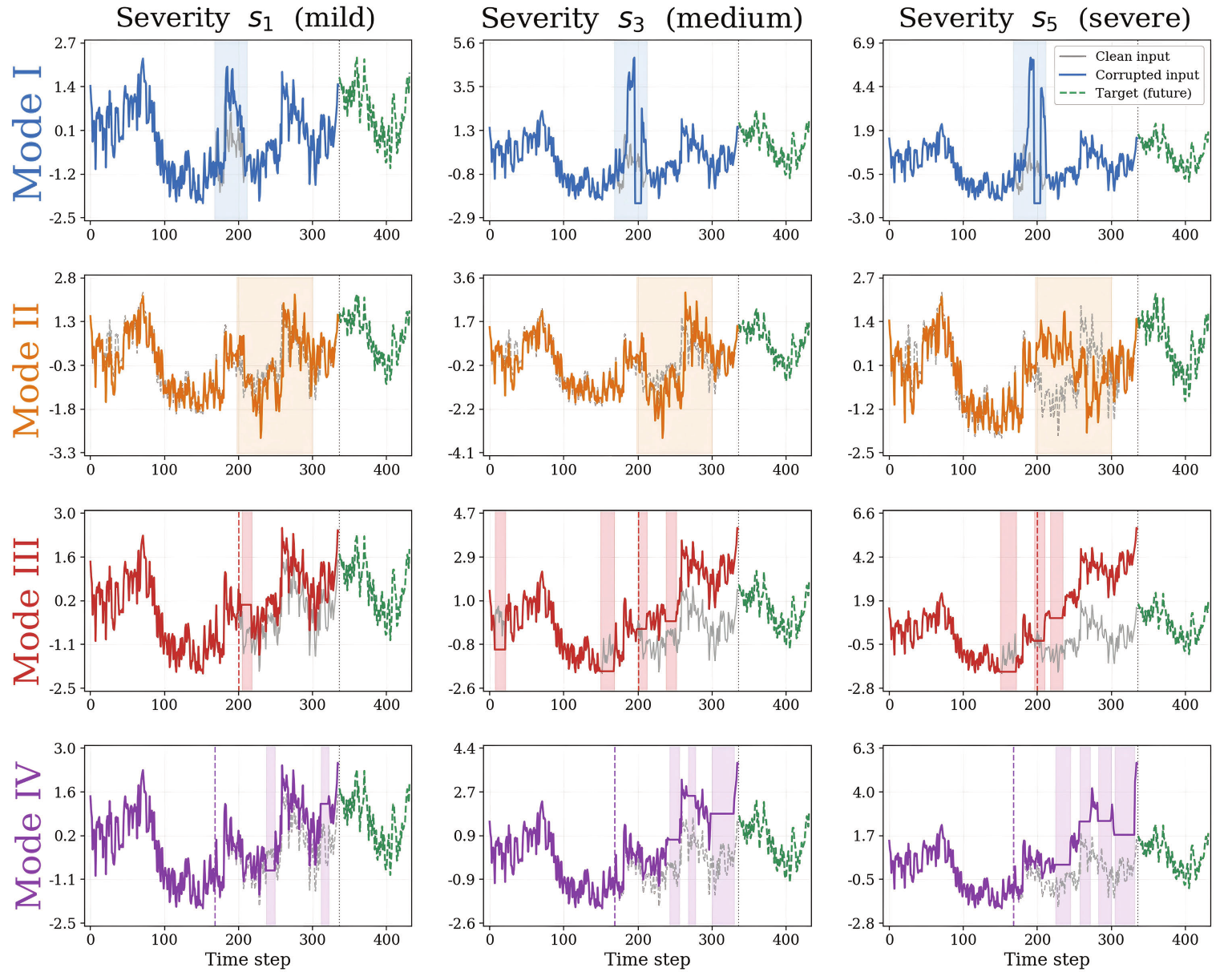}
\caption{Visual signatures of the four fault modes at three difficulties (clean in grey, faulted in color, untouched future target in green). 
}
\label{fig:waveforms}
\end{figure}

\section{The Four Fault Modes}
\label{sec:modes}
Drawing on documented failures in energy~\cite{busby2021cascading,andersson2005causes}, financial~\cite{forbes2002no}, and clinical~\cite{moor2021early,seymour2017time} systems, we organize meaningful TSF faults along two orthogonal axes (Figure~\ref{fig:taxonomy}). The first axis is scope: a fault either corrupts observations while the data-generating mechanism stays intact (observation-level, typically a transient event), or it alters the mechanism itself (mechanism-level, typically a persistent regime). The second axis is the variable scope: a fault either acts primarily on a single series (univariate) or on the cross-variable structure (multivariate). The four resulting modes (illustrated in Figure~\ref{fig:waveforms}) form a comprehensive evaluation.

\subsection{Mode~I: Time-Warped Shock (Observation $\times$ Univariate)}
\label{sec:modeI}
Mode~I is the observation-level, univariate case: a localized, shaped event whose timing is also distorted, as when a sensor briefly spikes or saturates and its reading is then displaced on the time axis by a buffering pipeline.
The fault therefore acts jointly in the value domain (an event with onset, peak, and decay) and on the temporal alignment.
This is a combination that i.i.d.\ noise, forbidding both temporal shape and a coherent time shift, cannot produce.

We select the critical window $W^\star$ with the importance score $S(W)$ (Eq.~\eqref{eq:window}), raising $\lambda_3,\lambda_4$ so the event lands on an internally dynamic window the model relies on, and inside it add a local event prototype at magnitude $\alpha$:
\begin{equation}
b_t=a_t\,m_t,\qquad \tilde{x}_t=x_t+\alpha\,b_t,\quad t\in W^\star,
\label{eq:m1event}
\end{equation}
where the support mask $m_t$ fixes which timesteps the event touches and the shape $a_t$ its local amplitude profile. The prototype is a narrow impulse (a transient spike), a decaying burst (a short overload), or a flat transient shift (a brief baseline drift). We then distort the event's position in time by re-sampling the shocked window through a temporal map $\phi(t)$, $x'_t=\tilde{x}_{\phi(t)}$ (by interpolation):
\begin{equation}
\phi(t)=
\begin{cases}
t+\Delta, & \text{shift}\\[1pt]
s+\alpha_w(t-s), & \text{scale}\\[1pt]
t+\Delta\sin\!\big(2\pi(t-s)/|W^\star|\big)+\xi_t, & \text{nonlinear}.
\end{cases}
\label{eq:m1warp}
\end{equation}
A shift offsets the timestamp, a scale stretches or compresses the event ($\alpha_w\!>\!1$ stretches, $\alpha_w\!<\!1$ compresses), and a nonlinear warp distorts its leading, middle, and trailing phases unevenly; only the corresponding segment is replaced, so the fault stays local. The scenario parameters $\Theta_{\mathrm{I}}=(\Theta_{\mathrm{evt}},\Theta_{\mathrm{wrp}},\Theta_{\mathrm{ctx}},\Theta_{\mathrm{cpl}})$ collect the event, the warp, the window context, and the event--warp coupling (the warp acting independently, centered on the event peak, or scaling with shock strength), and the difficulty $\delta_{\mathrm{I}}=\sum_k\beta_k D_k$ sums the matching event, warp, context, and coupling terms (Table~\ref{tab:difficulty}). \textit{Mode~I probes} whether a model over-relies on local peaks, short-term patterns, or precise temporal alignment.

\subsection{Mode~II: Dependency-Fracture Shock (Observation $\times$ Multivariate)}
\label{sec:modeII}
Mode~II is the observation-level, multivariate case: a group of variables that should move together has its lead--lag and gain structure covertly broken, where assets react to a common shock but with timing or sign that no longer matches their normal relationship. Each variate, viewed alone, still looks plausible.  Only the cross-variable structure is abnormal, which is a configuration that per-channel i.i.d.\ noise cannot represent.

Besides $W^\star$, we select the variable subset $S$ most embedded in the local dependency structure. Over the window neighborhood $N(W;r)$, we score each variable by its total lead--lag coupling to the rest,
\begin{equation}
\begin{gathered}
R_{ij}(W)=\max_{|\tau|\le\tau_{\max}}\big|\mathrm{Corr}\big(x^{(i)}_t,x^{(j)}_{t-\tau}\big)\big|,\\
G_i(W)=\sum_{j\ne i} R_{ij}(W),
\end{gathered}
\label{eq:m2leadlag}
\end{equation}
the maximum lagged cross-correlation summed over partners, and take $S=\mathrm{TopM}\{G_i(W)\}$. We inject a shared event prototype $u_t$ with heterogeneous per-variable gains ($\tilde{x}^{(i)}_t=x^{(i)}_t+g_i u_t$, $i\in S$), so the group appears to have undergone one common event but not identically; we then pick a root $r$ and, for each follower $j$, estimate its normal response template by least squares and deliberately falsify it:
\begin{equation}
\begin{gathered}
(\hat{\tau}_j,\hat{g}_j)=\arg\min_{\tau,g}\sum_{t\in N(W;r)}\big(x^{(j)}_t-g\,x^{(r)}_{t-\tau}\big)^2,\\
\tau'_j=\hat{\tau}_j+\Delta\tau_j,\quad g'_j=\hat{g}_j+\Delta g_j,\quad
{x'}^{(j)}_t=x^{(j)}_t+g'_j\,u_{t-\tau'_j},
\end{gathered}
\label{eq:m2fracture}
\end{equation}
where sign flips on $g'_j$ are permitted (turning a co-movement into an anti-movement) while the root keeps its normal response. The result looks like one shared event per channel, yet the inter-variable timing, strength, and direction are falsified. The parameters $\Theta_{\mathrm{II}}=(\Theta_{\mathrm{shk}},\Theta_{\mathrm{frc}},\Theta_{\mathrm{ctx}},\Theta_{\mathrm{scl}})$ collect the shock, the fracture offsets $(\Delta\tau,\Delta g)$, the window/variable context, and the participation scale, and $\delta_{\mathrm{II}}=\sum_k\beta_k D_k$ sums the shock, fracture, group, and position terms (Table~\ref{tab:difficulty}). \emph{Mode~II probes} whether a model that gains accuracy from cross-channel correlation.

\subsection{Mode~III: Regime-Transition Missingness (Mechanism $\times$ Univariate)}
\label{sec:modeIII}
Mode~III moves to the mechanism-level, univariate corner: the data-generating process itself changes, and samples go missing in a state-dependent, non-random way~\cite{little2019statistical}, such as when a grid enters emergency operation (new trend, period, and volatility)~\cite{busby2021cascading,ruan2020cross} while its monitoring drops blocks of samples around the very transition. The model therefore observes a biased, block-missing mixture of the old and new regimes, something neither value-level corruption nor random masking can produce.

We choose a switch center $\tau$ by a quality score that rewards a salient change in slope, period, or volatility near the forecast origin. For $t\ge\tau$ we rewrite the trend, season, and residual of $x_t=T_t+S_t+R_t$ into a new regime,
\begin{equation}
\tilde{T}_t=T_t+\Delta_\beta(t-\tau)+\Delta_b,\quad
\tilde{S}_t=a_s\,S_{\phi_s(t)},\quad
\tilde{R}_t=c_r R_t,
\label{eq:m3regime}
\end{equation}
where $(\Delta_\beta,\Delta_b)$ shift the trend slope and level, $a_s$ rescales the seasonal amplitude, $\phi_s(t)=\tau+\tfrac{P}{P'}(t-\tau)+\psi$ retimes its period and phase, and $c_r$ inflates the residual. We form the ideal new-regime trajectory $\bar{z}_t=\tilde{T}_t+\tilde{S}_t+\tilde{R}_t$ and blend it with the old one through a smooth sigmoid gate $\omega_t=\sigma((t-\tau)/w_\tau)$, $z_t=(1-\omega_t)x_t+\omega_t\bar{z}_t$, so the switch is a transition rather than a hard cut. The observation process then drops blocks where the system is most stressed, with missing-run start probability
\begin{equation}
p^{(i)}_{\mathrm{start}}(t)=\sigma\!\big(a_0+a_1 h_t+a_2 v^{(i)}_t +a_3 \varrho^{(i)}_t+a_4 c_i\big),
\label{eq:m3miss}
\end{equation}
where $h_t=4\omega_t(1-\omega_t)$ peaks at the switch core, $v^{(i)}_t$ is local volatility, $\varrho^{(i)}_t$ the standardized residual magnitude, and $c_i$ a per-channel fragility; run lengths are geometric, giving the observed series $x^{\mathrm{obs},(i)}_t=m^{(i)}_t z^{(i)}_t+(1-m^{(i)}_t)\,v^{\mathrm{fill},(i)}_t$. The model sees only this masked, regime-mixed history $\tilde{X}$ while the target stays clean ($\tilde{Y}=Y$); the parameters $\Theta_{\mathrm{III}}=(\Theta_{\mathrm{rgm}},\Theta_{\mathrm{mis}},\Theta_{\mathrm{ctx}},\Theta_{\mathrm{cpl}})$ and difficulty $\delta_{\mathrm{III}}=\sum_k\beta_k D_k$ (regime, missingness, proximity, coupling; Table~\ref{tab:difficulty}) follow accordingly. \emph{Mode~III probes} whether a model recovers the true future or instead extrapolates the spurious new regime and the gaps. It is the most structurally destructive mode as it rewrites the statistics a model sees in its input while hiding exactly the samples that would reveal the change.

\subsection{Mode~IV: Cascading Sensor-to-System Failure (Mechanism $\times$ Multivariate)}
\label{sec:modeIV}
Mode~IV is the remaining mechanism-level, multivariate case: a fault that originates at an upstream sensor, propagates along causal dependencies into downstream states, and finally degrades downstream observations~\cite{dobson2007complex,buldyrev2010catastrophic,nerc2011reliability,li2007analysis}. The fault itself propagates through the input window across three layers (root reading, downstream state, downstream observation).

Besides $W^\star$, we separate upstream drivers from downstream victims with a local directed-influence score
\begin{equation}
\Delta_{i\to j}(W)=\frac{\mathrm{Err}^{(j)}_{\mathrm{self}}-\mathrm{Err}^{(i\to j)}_{\mathrm{aug}}}{\mathrm{Err}^{(j)}_{\mathrm{self}}+\varepsilon}.
\label{eq:m4infl}
\end{equation}
The relative drop in $j$'s self-prediction error once $i$'s history is added as a predictor. The roots are the strongest drivers $R=\mathrm{Top}_{K_r}\{\sum_{j}\Delta_{i\to j}\}$ and the downstream set the strongest victims $D=\mathrm{Top}_{K_d}\{\sum_{r\in R}\Delta_{r\to j}\}$, taken disjoint from $R$, with trigger time $\tau_1$ where the roots deviate most from their window median. 
From $\tau_1$ on, each root stands for a faulted reading under one of four sensor-fault operators (\emph{bias drift}, \emph{saturation}, coarse \emph{quantization}, or \emph{stuck-at}), producing a fault error $e^{(r)}_t$ that seeds the cascade. The error reaches each downstream $j$ with gain $\gamma_{rj}$, delay $\Delta_{rj}$, and a decaying kernel $k_{rj}(\ell)=e^{-\ell/h_{rj}}/Z_{rj}$,
\begin{equation}
\zeta^{(j)}_t=\sum_{r\in R}\gamma_{rj}\sum_{\ell=0}^{n_k}k_{rj}(\ell)\, e^{(r)}_{t-\Delta_{rj}-\ell},
\label{eq:m4prop}
\end{equation}
dragging the downstream state to $z^{\prime(j)}_t=z^{(j)}_t+\zeta^{(j)}_t$, which in turn makes the downstream observation fragile and induces secondary dropouts whose rate grows with the local cascade magnitude $|\zeta^{(j)}_t|$. The observed input is therefore piecewise by channel role, faulted readings on $R$, displaced-and-masked values on $D$, clean elsewhere, while the target stays clean ($\tilde{Y}=Y$). The parameters $\Theta_{\mathrm{IV}}=(\Theta_{\mathrm{rt}},\Theta_{\mathrm{dwn}},\Theta_{\mathrm{flt}},\Theta_{\mathrm{prp}},\Theta_{\mathrm{sec}})$ and difficulty $\delta_{\mathrm{IV}}=\sum_k\beta_k D_k$ (root, cascade, delay, secondary; Table~\ref{tab:difficulty}) follow accordingly. \emph{Mode~IV probes} whether a model that aggregates across channels contains an upstream fault or instead spreads it into its own forecasts of the true future.

\begin{table}[]
\centering
\caption{Difficulty decomposition for the four fault modes. Each $\delta$ is the
weighted sum $\sum_k\beta_k D_k$ of the terms in its block.}
\label{tab:difficulty}
\footnotesize
\setlength{\tabcolsep}{2.2pt}
\renewcommand{\arraystretch}{1.22}
\begin{tabularx}{\columnwidth}{@{}
>{\raggedright\arraybackslash}p{0.27\columnwidth}
>{\raggedright\arraybackslash}X
>{\raggedright\arraybackslash}p{0.22\columnwidth}
@{}}
\toprule
\textbf{Term} & \textbf{Definition} & \textbf{Meaning} \\
\midrule

\rowcolor{black!6}
\multicolumn{3}{@{}l}{\textbf{Mode~I — Time-Warped Shock}} \\
$D_{\mathrm{event}}$    
& $\frac{\alpha}{\sigma_{\mathrm{loc}}}\frac{|\operatorname{supp}(m)|}{|W|}$ 
& energy density \\
$D_{\mathrm{warp}}$     
& $\frac{|\Delta|}{|W|}+|\alpha_w-1|+\kappa_{\mathrm{nl}}$ 
& time distortion \\
$D_{\mathrm{context}}$  
& $S_{\mathrm{pred}}(W)$ 
& predictive leverage \\
$D_{\mathrm{coupling}}$ 
& $c_{\mathrm{cpl}}$ 
& warp coupling \\

\rowcolor{black!6}
\multicolumn{3}{@{}l}{\textbf{Mode~II — Dependency Fracture}} \\
$D_{\mathrm{shock}}$    
& as $D_{\mathrm{event}}$ (on $u$) 
& shock energy \\
$D_{\mathrm{fracture}}$ 
& $\begin{aligned}[t]
   &\frac{1}{|S|-1}{\textstyle\sum_j}\Big(\frac{|\Delta\tau_j|}{|W|}+\frac{|\Delta g_j|}{|\hat g_j|+\varepsilon}\Big)
   \end{aligned}$ 
& template deviation \\
$D_{\mathrm{group}}$    
& $|S|/C$ 
& broken fraction \\
$D_{\mathrm{position}}$ 
& $e/L$ 
& edge proximity \\

\rowcolor{black!6}
\multicolumn{3}{@{}l}{\textbf{Mode~III — Regime Missingness}} \\
$D_{\mathrm{regime}}$   
& $\begin{aligned}[t]
   &\tfrac{|\Delta_\beta|}{\bar T'}
   +\big|\tfrac{\Delta_b}{\sigma_T}\big|
   +|a_s-1|\\
   &\quad
   +\big|\tfrac{P}{P'}-1\big|
   +\tfrac{|\psi|}{P}
   \end{aligned}$ 
& change size \\
$D_{\mathrm{missing}}$  
& $\bar p_{\mathrm{start}}\,\mathbb{E}[\ell_{\mathrm{run}}]$ 
& information loss \\
$D_{\mathrm{proximity}}$
& $\tau/L$ 
& switch proximity \\
$D_{\mathrm{coupling}}$ 
& $\operatorname{Corr}(\omega_t,\bar p_{\mathrm{start}})$ 
& switch--gap link \\

\rowcolor{black!6}
\multicolumn{3}{@{}l}{\textbf{Mode~IV — Cascading Failure}} \\
$D_{\mathrm{root}}$     
& $\frac{1}{|R|}\sum_{r}\mathbb{E}|e^{(r)}_t|/\sigma^{(r)}$ 
& fault magnitude \\
$D_{\mathrm{cascade}}$  
& $|D|\,\bar\gamma$ 
& reach $\times$ gain \\
$D_{\mathrm{delay}}$    
& $\operatorname{Var}\{\Delta_{rj}\}$ 
& delay spread \\
$D_{\mathrm{secondary}}$
& $\frac{1}{|D|}\sum_{j}\bar p^{(j)}_{\mathrm{start}}$ 
& dropout rate \\

\bottomrule
\end{tabularx}
\end{table}

\subsection{difficulty Control and Fault Composition}
\label{sec:difficulty-detail}
Each mode's difficulty $\delta=\kappa(\Theta)$ is a weighted sum of four interpretable terms, $\delta=\sum_k\beta_k D_k$, so a hard instance records not only that it is hard but why. Table~\ref{tab:difficulty} states the terms explicitly.
We also introduce difficulty to systematically conduct the evaluation.
The five levels $s_1\!\to\!s_5$ sweep $\Theta$ along $\kappa$ so that every term increases monotonically. Further, because $\kappa$ is monotone and each term is interpretable, the resulting robustness curves (Sec.~\ref{sec:rq5}) could measure the sensitivity to a named parameter.

The four modes are operators and therefore composable. A compound condition is the composition
\begin{equation}
\mathcal{T}_{\Theta}\;=\;\mathcal{T}_{\Theta_k}\circ\cdots\circ \mathcal{T}_{\Theta_1},
\label{eq:compose}
\end{equation}
which is generally non-commutative. A drift followed by an impulse yields a different faulted window than the reverse. Thus, the order is itself a scenario parameter. Because each operator exposes its own $\Theta$, the contribution of every fault in a composite instance remains attributable, turning ``which failure degraded this model'' from a guess into a decomposition. We treat single-mode evaluation as the core protocol and leave systematic compositional sweeps to future work.

\section{TS-Fault Benchmark and Experimental Setup}
\label{sec:setup}
With the framework (Sec.~\ref{sec:framework}) and the four modes (Sec.~\ref{sec:modes}) in place, we now instantiate TS-Fault as a concrete benchmark and describe the protocol of our large-scale evaluation study, including the datasets, the evaluated models, the paired evaluation procedure, the metrics, and the reproducibility details.

\begin{table}[]
\centering
\caption{The six multivariate datasets in TS-Fault.}
\label{tab:datasets}
\footnotesize
\renewcommand{\arraystretch}{1.15}
\setlength{\tabcolsep}{4pt}
\begin{tabular*}{\columnwidth}{@{\extracolsep{\fill}}llrlr@{}}
\toprule
\textbf{Dataset} & \textbf{Domain} & \textbf{\#Ch.} & \textbf{Granularity} & \textbf{Length} \\
\midrule
ETTh1        & Energy  & 7   & 1\,hour  & 17{,}420 \\
ETTh2        & Energy  & 7   & 1\,hour  & 17{,}420 \\
ETTm1        & Energy  & 7   & 15\,min  & 69{,}680 \\
ETTm2        & Energy  & 7   & 15\,min  & 69{,}680 \\
Electricity  & Load    & 321 & 1\,hour  & 26{,}304 \\
Weather      & Climate & 21  & 10\,min  & 52{,}696 \\
\bottomrule
\end{tabular*}
\end{table}

\subsection{Datasets}
\label{sec:datasets}
We build TS-Fault on six widely used multivariate long-term forecasting datasets spanning the energy, load, and climate domains (Table~\ref{tab:datasets})~\cite{zhou2021informer,wu2021autoformer,lai2018modeling}. We chose them to vary widely in the two properties that interact with our fault modes: dimensionality (from $7$ channels in ETT to $321$ in Electricity) and sampling granularity (from $10$-minute to hourly). 
Dimensionality matters because Mode~II and Mode~IV act on cross-variable structure. For example, the $321$ densely correlated channels of Electricity make it the most demanding test of cross-channel fracture and propagation. 
Following standard forecasting protocol in~\cite{chen2023tsmixer,nie2022time,zhang2024multi}, each clean test window comprises a length-$336$ history and a length-$96$ target.

\begin{figure}[t]
\centering
\includegraphics[width=\columnwidth]{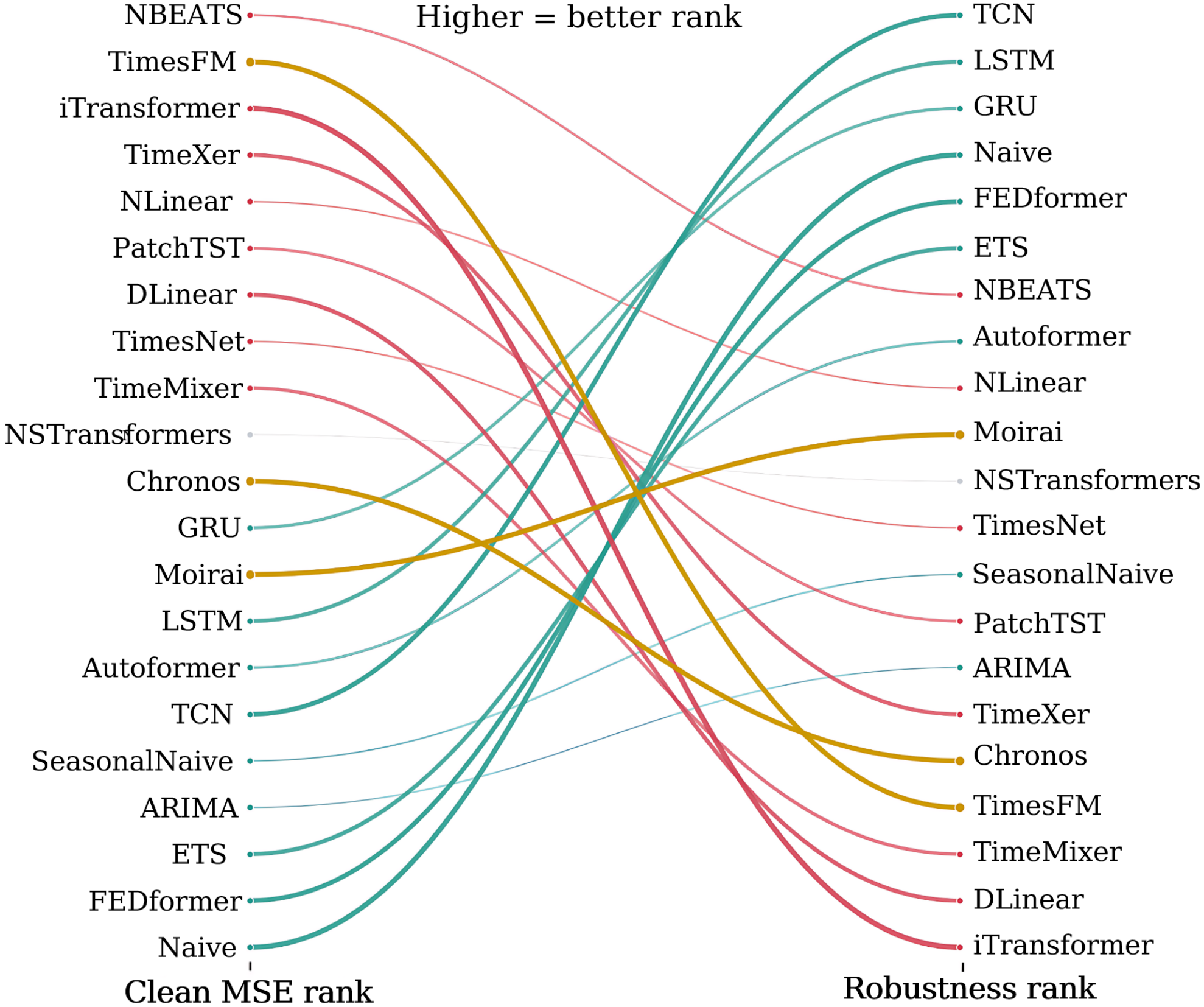}
\caption{Each line links a model's clean-accuracy rank (left) to its robustness rank (right); 
}
\label{fig:nfl}
\end{figure}

\begin{table*}[t]
\centering
\caption{Aggregate results of the 21 models, grouped by model category, over the metrics defined in Sec.~\ref{sec:metrics}. Bold marks the best value per column ($k=10^{3}$).}
\label{tab:overall}
\scriptsize
\setlength{\tabcolsep}{4pt}
\resizebox{\textwidth}{!}{%
\begin{tabular}{@{}l rr rr rr rrrr r rrr@{}}
\toprule
\textbf{Model}
& \multicolumn{2}{c}{\textbf{Clean}}
& \multicolumn{2}{c}{\textbf{Faulted}}
& \multicolumn{2}{c}{\textbf{Robustness}}
& \multicolumn{4}{c}{\textbf{Per-mode RD (\%)}}
& \textbf{}
& \multicolumn{3}{c}{\textbf{Rank}} \\
\cmidrule(lr){2-3}\cmidrule(lr){4-5}\cmidrule(lr){6-7}\cmidrule(lr){8-11}\cmidrule(lr){13-15}
& MSE & MAE & MSE & MAE & $\Delta$MSE & $r$ & I & II & III & IV & $d_{10}/d_{02}$ & $R_{\mathrm{cln}}$ & $R_{\mathrm{rob}}$ & $\Delta$rank \\
\midrule

\rowcolor{black!6}
\multicolumn{15}{@{}l}{\textbf{Statistical}} \\
Naive~\cite{hyndman2018forecasting}                 & 1.238 & 0.780 & 129.5 & 4.798 & 128.3 & 105.7 & 0 & 0 & 27.4k & 2.1k & 23.2 & 21 & 4 & -17 \\
SeasonalNaive~\cite{hyndman2018forecasting}         & 0.905 & 0.620 & 118.5 & 4.469 & 117.6 & 165.5 & 11 & 0 & 26.8k & 2.6k & 24.1 & 17 & 13 & -4 \\
ARIMA~\cite{box2015time}                 & 0.999 & 0.694 & 121.7 & 4.531 & 120.7 & 120.7 & -1 & 0 & 27.8k & 4.8k & 25.7 & 18 & 15 & -3 \\
ETS~\cite{hyndman2008forecasting}                   & 1.044 & 0.730 & 126.7 & 4.708 & 125.7 & 122.9 & 0 & 0 & 28.2k & 2.4k & 23.5 & 19 & 6 & -13 \\
\midrule

\rowcolor{black!6}
\multicolumn{15}{@{}l}{\textbf{Linear / lightweight}} \\
DLinear~\cite{zeng2023dlinear}               & 0.562 & 0.506 & 66.85 & 3.338 & 66.29 & 197.3 & 7 & 1 & 25.4k & 1.8k & 23.5 & 7 & 20 & +13 \\
NLinear~\cite{zeng2023dlinear}               & 0.540 & 0.521 & 72.23 & 3.397 & 71.69 & 142.1 & 15 & 1 & 27.3k & 2.5k & 23.7 & 5 & 9 & +4 \\
N-BEATS~\cite{Oreshkin2020N-BEATS:}               & \textbf{0.449} & 0.458 & 21.59 & 1.670 & 21.14 & 54.6 & 6 & 0 & 15.3k & 465 & 20.1 & \textbf{1} & 7 & +6 \\
\midrule

\rowcolor{black!6}
\multicolumn{15}{@{}l}{\textbf{Recurrent / conv.}} \\
LSTM~\cite{hochreiter1997long}                  & 0.733 & 0.638 & 0.780 & 0.661 & \textbf{0.047} & \textbf{1.07} & 0 & 0 & \textbf{9} & \textbf{8} & \textbf{1.0} & 14 & 2 & -12 \\
GRU~\cite{cho2014learning}                   & 0.680 & 0.607 & \textbf{0.775} & \textbf{0.651} & 0.095 & 1.14 & 0 & 0 & 26 & 23 & \textbf{1.0} & 12 & 3 & -9 \\
TCN~\cite{bai2018empirical}                   & 0.874 & 0.710 & 7.389 & 1.300 & 6.515 & 7.90 & 0 & 0 & 900 & 57 & 8.1 & 16 & \textbf{1} & -15 \\
\midrule

\rowcolor{black!6}
\multicolumn{15}{@{}l}{\textbf{Decomposition Transf.}} \\
Autoformer~\cite{wu2021autoformer}            & 0.822 & 0.658 & 24.19 & 2.225 & 23.37 & 48.0 & 4 & 0 & 6.6k & 388 & 16.4 & 15 & 8 & -7 \\
FEDformer~\cite{zhou2022fedformer}             & 1.072 & 0.793 & 24.29 & 2.311 & 23.22 & 22.6 & 3 & 0 & 5.1k & 270 & 15.6 & 20 & 5 & -15 \\
\midrule

\rowcolor{black!6}
\multicolumn{15}{@{}l}{\textbf{Attention / SOTA}} \\
PatchTST~\cite{nie2022time}              & 0.551 & 0.500 & 86.63 & 3.755 & 86.08 & 262.6 & 13 & 1 & 37.3k & 2.9k & 24.2 & 6 & 14 & +8 \\
iTransformer~\cite{liu2024itransformer}          & 0.529 & 0.495 & 83.16 & 3.668 & 82.63 & 272.3 & 11 & 2 & 38.0k & 3.4k & 24.4 & 3 & 21 & +18 \\
TimeXer~\cite{wang2024timexer}               & 0.537 & 0.495 & 87.98 & 3.774 & 87.44 & 301.2 & 9 & 0 & 39.7k & 3.1k & 24.0 & 4 & 16 & +12 \\
TimeMixer~\cite{wang2024timemixer}             & 0.590 & 0.518 & 90.08 & 3.842 & 89.49 & 264.2 & 7 & 1 & 38.9k & 2.8k & 24.6 & 9 & 19 & +10 \\
TimesNet~\cite{wu2022timesnet}              & 0.571 & 0.528 & 53.34 & 2.882 & 52.77 & 141.7 & 8 & 1 & 19.9k & 1.6k & 22.5 & 8 & 12 & +4 \\
Nonstationary Transf.r~\cite{liu2022non} & 0.608 & 0.550 & 55.16 & 2.928 & 54.55 & 101.1 & 18 & 3 & 20.0k & 1.9k & 24.4 & 10 & 11 & +1 \\
\midrule

\rowcolor{black!6}
\multicolumn{15}{@{}l}{\textbf{Foundation, zero-shot}} \\
TimesFM~\cite{das2023decoder}               & 0.516 & \textbf{0.436} & 162.7 & 4.887 & 162.2 & 555.2 & 8 & 1 & 87.2k & 10.2k & 27.9 & 2 & 18 & +16 \\
Chronos~\cite{ansari2024chronos}               & 0.613 & 0.499 & 165.6 & 5.068 & 165.0 & 512.5 & 3 & 0 & 69.0k & 5.8k & 25.5 & 11 & 17 & +6 \\
Moirai~\cite{woo2024unified}                & 0.682 & 0.538 & 153.1 & 4.954 & 152.4 & 365.6 & 2 & 0 & 45.8k & 4.9k & 25.7 & 13 & 10 & -3 \\

\bottomrule
\end{tabular}%
}
\end{table*}

\subsection{Evaluated Models}
\label{sec:models}
We evaluate $21$ forecasting models grouped into six methodological categories
(Table~\ref{tab:overall}), from classical statistics to pretrained foundation models. These categories cover fundamentally different mechanisms: autoregressive and exponential-smoothing statistics; linear and basis-expansion maps; recurrent and convolutional sequence models; decomposition-based and modern attention Transformers; and the more recent trend of large time series foundation models. 
The first five categories are fit to each dataset under a common configuration (Sec.~\ref{sec:protocol}), that is, the deep models trained with a shared optimizer recipe and the statistical models fit per window. For the time series foundation models, they are evaluated strictly zero-shot with their released pretrained weights and are never fine-tuned on our data, so their performance (both clean and faulted situations) reflects the general capability after their pretraining process.


\subsection{Protocol}
\label{sec:protocol}

\noindent\textbf{Training.} Trained models follow one shared recipe to keep comparisons fair: at most $10$ epochs with early stopping (patience $3$), the Adam optimizer, learning rate $10^{-4}$, MSE loss, and batch size $32$ (reduced to $16$ on the $321$-channel Electricity). Statistical models are fit directly on each window and foundation models perform zero-shot inference.

\noindent\textbf{Difficulty.} Each mode is instantiated at five increasing difficulty levels $s_1$ (mildest) to $s_5$ (most severe), obtained by sweeping the mode's scenario parameters $\Theta$ along the difficulty map $\kappa$ (Sec.~\ref{sec:difficulty}). These correspond to the released configurations $d_{02},d_{04},d_{06},d_{08},d_{10}$.

\noindent\textbf{Paired clean/corrupt evaluation.} The benchmark is organized into $6$ datasets $\times$ $4$ modes $\times$ $5$ difficulties $=120$ configurations, each including $20$ paired windows. A pair is a clean context $X$ and its faulted counterpart $\tilde{X}=\mathcal{T}_\Theta(X)$, both forecast by the same model and compared against the same target. 
This pairing separates the two quantities we focus on. The clean error measures a model's intrinsic accuracy, while the increase in error from clean to faulted input measures its sensitivity to faults.

\subsection{Metrics}
\label{sec:metrics}
For a model $f$ and a fault mode $\mathcal{F}_i$, let $L(f)=\mathbb{E}[\ell(f(X),Y)]$ be its clean error on the non-corrupted data and $\mathrm{AVG}(f,\mathcal{F}_i)$ the average faulted error (Eq.~\eqref{eq:avg}).
We use the standard squared-error loss to evaluate the forecasting accuracy~\cite{hyndman2006another,gneiting2007strictly}. 
We also introduce and report the absolute degradation $\Delta\mathrm{MSE}_i(f)=\mathrm{AVG}(f,\mathcal{F}_i)-L(f)$, 
the robustness ratio $r_i(f)=\mathrm{AVG}(f,\mathcal{F}_i)/L(f)$, 
and the relative degradation $\mathrm{RD}_i(f)=\big(r_i(f)-1\big)\times100\%$. 
A ratio $r_i=1$ denotes perfect robustness. 
We treat any cell with $r_i\ge 10$ as a catastrophic failure: a full order-of-magnitude error inflation, large enough to flip a downstream operational decision. 
A model's aggregated ratio averages its within-dataset ratio across the six datasets, so that high-magnitude datasets do not dominate the summary. 
Because $r$ is a ratio, it is sensitive to the scale of the clean error.
A model with a very small $L(f)$ can post a large $r$ from only a modest absolute change, so two models with nearly identical $\Delta\mathrm{MSE}$ can differ largely in $r$ (e.g.\ N-BEATS vs.\ FEDformer in Table~\ref{tab:overall}). We therefore report $\Delta\mathrm{MSE}$, $r$, and $\mathrm{RD}$ side by side.

The central metric is the Spearman rank correlation $\rho$ between a model's clean-accuracy ranking and its faulted (robustness) ranking, computed globally and per mode. The accompanying two-sided $p$-value tests the null hypothesis of no rank association ($\rho=0$); if $p<0.05$, the correlation is regarded as statistically significant, meaning that the observed preservation or reversal of rankings is unlikely to arise from random ordering alone. 
Here $\rho\!\to\!1$ means the clean leaderboard predicts deployed robustness, whereas $\rho\!\to\!0$ means it does not, which directly quantifies the model-selection risk that motivates this work.
To reduce sensitivity to a few diverging configurations, clean rankings are formed from the mean clean error and robustness rankings from the median relative degradation $\mathrm{RD}_i$.
The mean faulted error is dominated by a handful of extreme configurations, so we report the median which is the more faithful summary of typical robustness.
Unlike $r$, the median over configurations is also robust to the few inflated ratios a small clean error can produce.

\subsection{Reproducibility}
\label{sec:repro}
TS-Fault is designed to be fully reproducible. We release: (i) the parameterized fault generators for all four modes, together with their scenario-parameter schemas $\Theta$ and the unified window-importance mechanism; (ii) the generated benchmark instances, \textit{i.e.}, paired clean/corrupt windows with their targets and exposed $\Theta$ and difficulty $\delta$; (iii) the evaluation and per-model configurations; and (iv) trained checkpoints. The codes are available here: \href{https://github.com/Ray-zyy/TS-Fault}{github.com/Ray-zyy/TS-Fault}.
Because faulted instances are produced by an explicit operator at evaluation time, the benchmark can be regenerated at any difficulty, and previously unexposed $\Theta$ combinations can be held out to guard against benchmark overfitting in the future.

\section{Results and Analysis}
\label{sec:results}

We organize our findings around six research questions, moving from the headline model-selection result (RQ1) to the validity checks on difficulty and data dependence (RQ5--RQ6). 

\noindent\textbf{Overall Performance Summary.} Table~\ref{tab:overall} summarizes results for all 21 models, reporting clean and faulted forecasting performance alongside key robustness metrics, including degradation, robustness ratio, per-mode degradation, difficulty slope, and rank shifts. Table~\ref{tab:overall} highlights three findings. \textbf{First}, the $R_{\mathrm{cln}}$ and $R_{\mathrm{rob}}$ columns are near mirror images: the most robust models, GRU, LSTM, and TCN, are only mid-pack on clean-data accuracy, whereas the clean-accuracy leaders (N-BEATS, TimesFM, iTransformer, TimeXer) sink to the bottom of the robustness order. \textbf{Second}, the per-mode columns are negligible under the observation-level Modes~I/II ($\le\!18\%$) yet explode to tens of thousands of percent under the mechanism-level Modes~III/IV. \textbf{Third}, the three foundation models (the bottom three rows) pair the best clean-data accuracy in the table with the worst value in every robustness column.

\subsection{RQ1: Does clean-data accuracy Predict Robustness?}
\label{sec:rq1} 
No. The Spearman correlation between a model's clean-accuracy rank and its robustness rank is negative and significant, $\rho=-0.544$ ($p=0.011$) across all $21$ models; restricted to the $18$ non-foundation models it is $-0.509$ ($p=0.031$), so adding pretrained models strengthens rather than dilutes the effect. The dislocations are large: iTransformer, the third-most-accurate model on clean data, ranks dead last ($21$st) in robustness, a shift of $+18$ ranks, the largest in the study, and the foundation model TimesFM, second on clean-data accuracy, falls to $18$th ($+16$). Conversely, TCN, LSTM, and GRU (clean-accuracy ranks $16$/$14$/$12$) occupy the top three robustness ranks.

Figure~\ref{fig:nfl} traces these crossings, and the $R_{\mathrm{cln}}$, $R_{\mathrm{rob}}$, and $\Delta\mathrm{rank}$ columns of Table~\ref{tab:overall} list them numerically. These robustness ranks use the median relative degradation (Sec.~\ref{sec:metrics}); this differs from a ranking by mean faulted MSE, under which the foundation models, whose few extreme configurations inflate the mean, rank worst instead. Both orderings deliver the same verdict: clean-data accuracy does not predict robustness. The practical consequence is the model-selection risk that motivates this work: choosing a forecaster by clean MSE, today's default, systematically favors the architecture most likely to fail once structured faults appear in deployment.

\subsection{RQ2: Is Failure Impact Uniform or Stratified?}
\label{sec:rq2}
\begin{figure}[]
\centering
\includegraphics[width=\columnwidth]{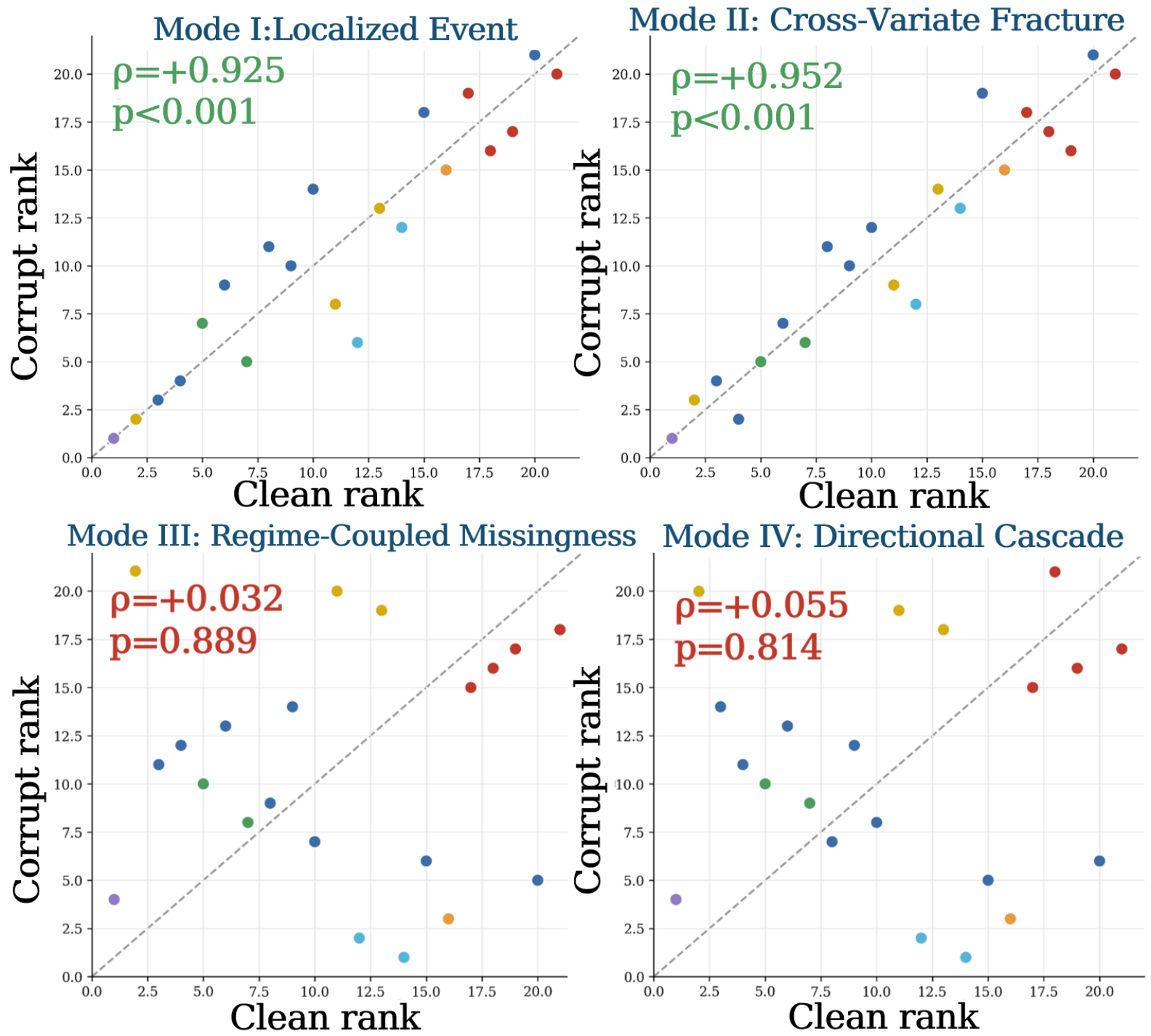}
\caption{Clean rank vs.\ faulted rank, per mode. The dashed diagonal denotes a perfectly preserved ranking. Modes~I/II hug the diagonal ($\rho>0.92$); Modes~III/IV scatter almost completely ($\rho\approx0.03$/$0.06$). Gold points are foundation models, concentrated in the high-fragility region under Modes~III/IV.}
\label{fig:rankcorr}
\end{figure}

Sharply stratified, and this stratification is the methodological payoff of TS-Fault. Table~\ref{tab:rankcorr} reports the clean-vs-faulted rank correlation per mode. Under the two observation-level modes, the clean ranking is almost perfectly preserved (Mode~I $\rho=0.925$, Mode~II $\rho=0.952$, both $p<0.001$); under the two mechanism-level modes, it is essentially obliterated (Mode~III $\rho=0.032$, $p=0.889$; Mode~IV $\rho=0.055$, $p=0.814$).Figure~\ref{fig:rankcorr} shows the same split visually: the Mode~I/II panels hug the identity diagonal, while the Mode~III/IV panels scatter at random. The same stratification is already visible in Table~\ref{tab:overall}: the Mode~I and Mode~II columns stay below $18\%$ for every model, whereas the Mode~III and Mode~IV columns run from hundreds to tens of thousands of percent. Adding the foundation models again pushes the mechanism-level correlations further toward zero (from $0.21$/$0.19$ at $18$ models to $0.03$/$0.06$). The reading is direct and actionable. Under observation-level faults, a localized event or a covertly fractured dependency, clean MSE remains a sound selection proxy. Under mechanism-level faults, a regime switch coupled with missingness, or a fault cascading along a sensing chain, clean MSE carries essentially no information about which model will survive. Because TS-Fault exposes which mode produced a degradation, it can attribute fragility to a named mechanism rather than reporting an undifferentiated aggregate. Table~\ref{tab:permode} lists the per-mode median degradation for all $21$ models, sorted by Mode~III: Modes~I/II remain $\le\!18\%$ for every architecture, whereas Mode~III ranges from $9\%$ (LSTM) to $87{,}000\%$ (TimesFM) and Mode~IV from $8\%$ to $10{,}000\%$, a three-to-four-order-of-magnitude gap between the two halves of the taxonomy.

\begin{table}[t]
\centering
\caption{Median relative degradation (\%) by fault mode for all $21$ models, sorted by Mode~III ($k=10^{3}$). Observation-level Modes~I/II stay $\le\!18\%$ for every architecture; mechanism-level Modes~III/IV explode by three to four orders of magnitude. Foundation rows shaded.}
\label{tab:permode}
\footnotesize
\setlength{\tabcolsep}{6pt}
\renewcommand{\arraystretch}{1.05}
\begin{tabular*}{\columnwidth}{@{\extracolsep{\fill}}l rrrr@{}}
\toprule
\textbf{Model} & \textbf{Mode~I} & \textbf{Mode~II} & \textbf{Mode~III} & \textbf{Mode~IV} \\
\midrule
LSTM           & 0  & 0 & 9   & 8   \\
GRU            & 0  & 0 & 26  & 23  \\
TCN            & 0  & 0 & 900 & 57  \\
FEDformer      & 3  & 0 & 5k  & 270 \\
Autoformer     & 4  & 0 & 7k  & 388 \\
N-BEATS        & 6  & 0 & 15k & 465 \\
TimesNet       & 8  & 1 & 20k & 2k  \\
NSTransformers & 18 & 3 & 20k & 2k  \\
DLinear        & 7  & 1 & 25k & 2k  \\
SeasonalNaive  & 11 & 0 & 27k & 3k  \\
NLinear        & 15 & 1 & 27k & 3k  \\
Naive          & 0  & 0 & 27k & 2k  \\
ARIMA          & -1 & 0 & 28k & 5k  \\
ETS            & 0  & 0 & 28k & 2k  \\
PatchTST       & 13 & 1 & 37k & 3k  \\
iTransformer   & 11 & 2 & 38k & 3k  \\
TimeMixer      & 7  & 1 & 39k & 3k  \\
TimeXer        & 9  & 0 & 40k & 3k  \\
Moirai  & 2 & 0 & 46k & 5k  \\
Chronos & 3 & 0 & 69k & 6k  \\
TimesFM & 8 & 1 & 87k & 10k \\
\bottomrule
\end{tabular*}
\end{table}

\begin{table}[]
\centering
\caption{Spearman $\rho$ between clean and faulted model rankings, per mode. Observation-level faults preserve the ranking; mechanism-level faults destroy it.}
\label{tab:rankcorr}
\footnotesize
\renewcommand{\arraystretch}{1.08}
\setlength{\tabcolsep}{3pt}
\begin{tabular*}{\columnwidth}{@{\extracolsep{\fill}} l c c l @{}}
\toprule
\textbf{Mode} & $\rho$ & $p$ & \textbf{Effect} \\
\midrule
I~~(Time-Warped Shock)    & $+0.925$ & $<0.001$ & ranking preserved \\
II~~(Dependency Fracture) & $+0.952$ & $<0.001$ & ranking preserved \\
III~(Regime Missingness)  & $+0.032$ & $0.889$  & ranking destroyed \\
IV~(Cascading Failure)    & $+0.055$ & $0.814$  & ranking destroyed \\
\bottomrule
\end{tabular*}
\end{table}

\subsection{RQ3: Where Do Catastrophic Failures Concentrate?}
\label{sec:rq3} 
\begin{figure}[]
\centering
\includegraphics[width=\columnwidth]{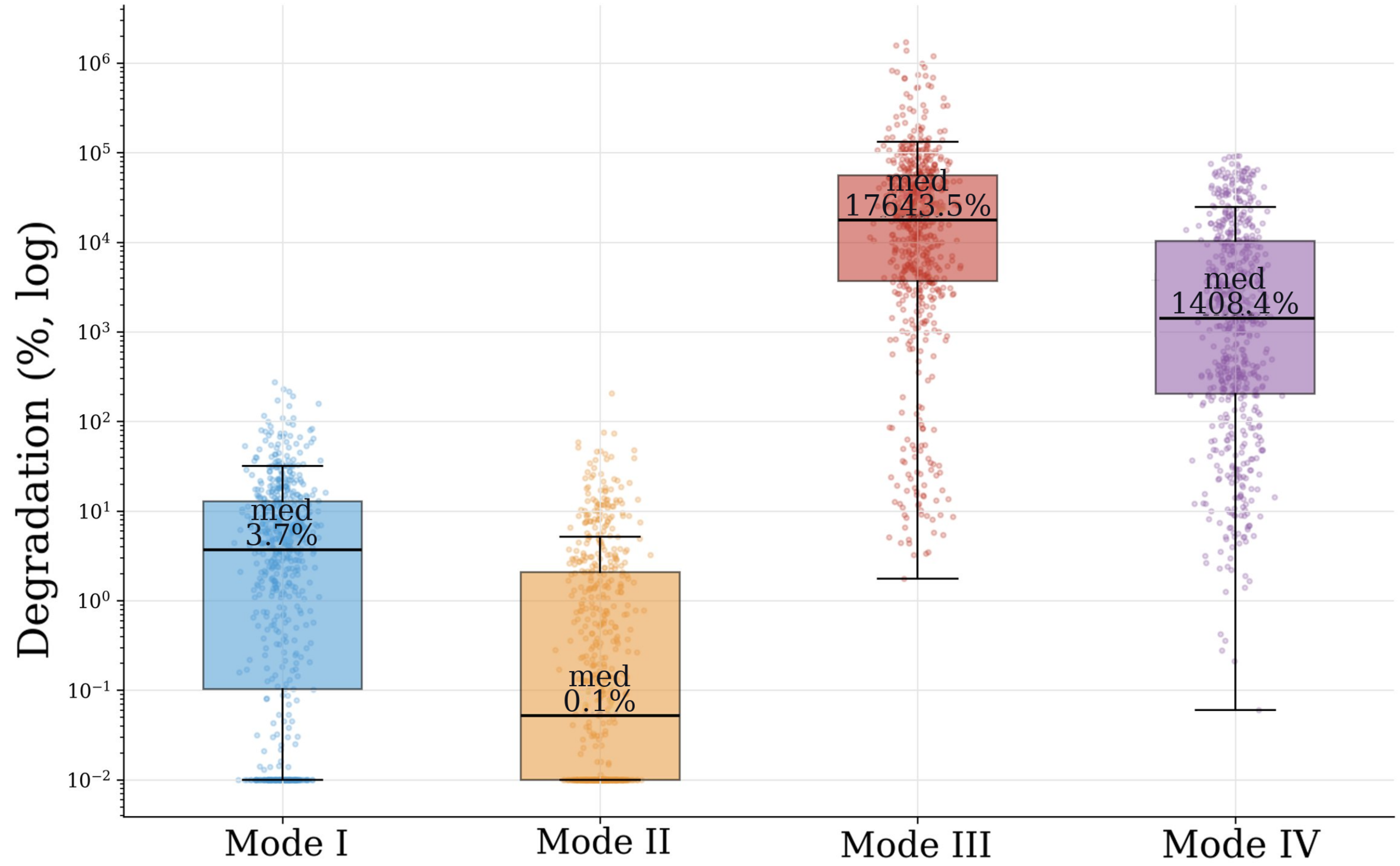}
\caption{Per-mode degradation distributions (log $y$; box = IQR, points = single configuration). Modes~I/II are low and tightly concentrated; Modes~III/IV are higher by several orders of magnitude and far more dispersed.}
\label{fig:degdist}
\end{figure}

Entirely in the mechanism-level modes. Counting every cell whose error inflates by at least an order of magnitude ($r\ge10$), we find $884$ catastrophic failures across the grid, distributed with striking asymmetry: $0$ in Mode~I, $0$ in Mode~II, $537$ in Mode~III ($85.9\%$ of its $625$ configurations), and $347$ in Mode~IV ($55.5\%$), so Modes~III/IV account for $100\%$ of all catastrophic failures. Figure~\ref{fig:degdist} explains why: the observation-level modes degrade by a median of $\approx\!3.7\%$ (Mode~I) and $\approx\!0.1\%$ (Mode~II) with tight spread, whereas the mechanism-level modes degrade by medians of $\approx\!1.8\times10^{4}\%$ (Mode~III) and $\approx\!1.4\times10^{3}\%$ (Mode~IV) and are enormously dispersed, a single model can degrade by $100\%$ or by $100{,}000\%$ depending on the configuration. That dispersion is itself a risk: a model's behavior under mechanism-level faults cannot be estimated in advance. By model, the catastrophic count is led by TimesFM ($53$) and the attention/foundation cluster (TimeXer, iTransformer, TimeMixer, Chronos, Moirai, $52$ each), against only $17$ for TCN.

\subsection{RQ4: Are Pretrained Foundation Models More Robust?}
\label{sec:rq4}
\begin{figure}[]
\centering
\includegraphics[width=\columnwidth]{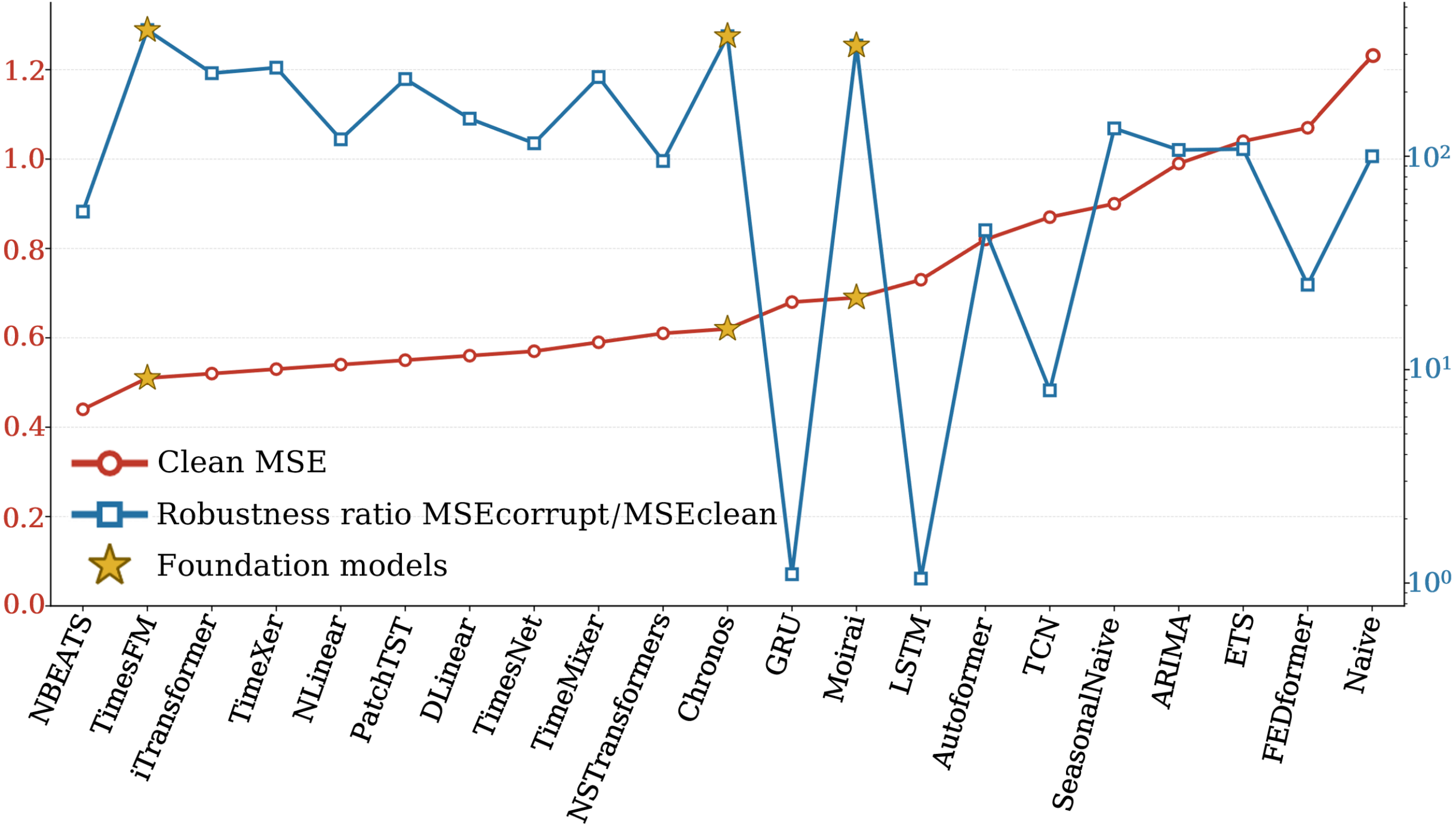}
\caption{Clean MSE and robustness ratio across the $21$ models. Gold stars mark foundation models.}
\label{fig:foundation}
\end{figure}

They are the opposite, strong but fragile. On clean data, the three foundation models are top-tier: TimesFM attains MSE $0.516$ (second overall) and MAE $0.436$ (first), with Chronos ($0.613$) and Moirai ($0.682$) ahead of many specialists. Yet their robustness ratios, $555$/$512$/$366$, are the three worst of all $21$ models, and their mean faulted errors ($163$/$166$/$153$) are the three highest. The verdict does not hinge on the ratio: in \emph{absolute} terms their degradation $\Delta\mathrm{MSE}$ ($162$/$165$/$152$) is also the largest in the table, so the fragility is not an artifact of dividing by a small clean error.
The gap is widest exactly where it matters: under Mode~III the median degradation reaches $87{,}000\%$ for TimesFM, $69{,}000\%$ for Chronos, and $46{,}000\%$ for Moirai, against $40{,}000\%$ for the worst specialist (TimeXer); on the $321$-channel Electricity dataset TimesFM's ratio peaks at $2090$. Figure~\ref{fig:foundation} portrays the two faces side by side. A plausible mechanism is that zero-shot models rely on strong learned priors~\cite{geirhos2020shortcut,d2022underspecification}, typical periodicity, smoothness, cross-channel co-movement, and, lacking any online adaptation~\cite{kim2021reversible} to the current series, cannot recover when Modes~III/IV rewrite those very structures; the prior that buys clean-data accuracy then amplifies the error. Pretraining thus delivers accuracy, not fault robustness, and including these models is precisely what strengthens the global No-Free-Lunch correlation from $-0.509$ to $-0.544$.

\subsection{RQ5: Is Difficulty Controllable and Monotonic?}
\label{sec:rq5} 
\begin{figure}[]
\centering
\includegraphics[width=.95\columnwidth]{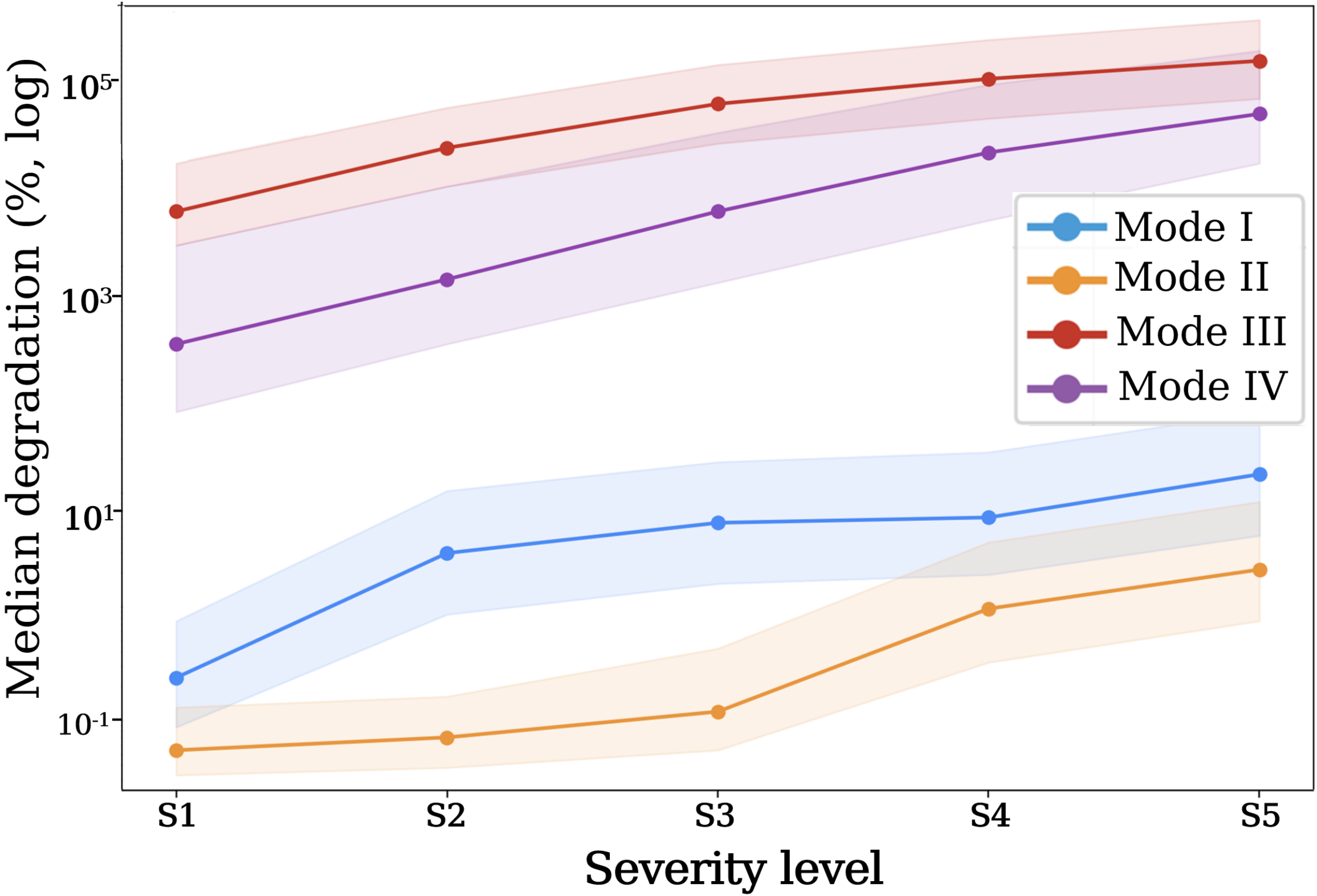}
\caption{Degradation grows monotonically with difficulty for all four modes (median over $21$ models; log $y$, shaded IQR). Mechanism-level Modes~III/IV dominate observation-level Modes~I/II by orders of magnitude at every level.}
\label{fig:difficulty}
\end{figure}

Yes. Degradation rises monotonically from difficulty $s_1$ to $s_5$ for every mode (Figure~\ref{fig:difficulty}), confirming that the difficulty map $\kappa$ behaves as designed and supports graded robustness curves rather than single-point comparisons. The slope $d_{10}/d_{02}$ also separates three clear sensitivity tiers: low ($\approx\!1.0$) for GRU and LSTM; mid ($\approx\!8$--$20$) for TCN, FEDformer, Autoformer, and N-BEATS; and high ($\approx\!22$--$28$) for the attention SOTA, the linear models (DLinear, NLinear), the statistical models, and all three foundation models, with TimesFM the steepest at $27.9$. The tiers track architecture: recurrent hidden-state smoothing dilutes an injected fault, trend--season decomposition is largely immune to aggregate statistic change, whereas per-position attention and frozen zero-shot priors transmit difficulty into error almost linearly. Table~\ref{tab:difflevels} reports the per-level mean error for all $21$ models (averaged over the six datasets and four modes): error grows with severity across essentially every row, and the slope $d_{10}/d_{02}$ separates the same three tiers, GRU and LSTM flat at $\approx\!1.0$, TCN and the decomposition models in the middle, and the attention SOTA, linear, statistical, and all three foundation models climbing $\sim\!22$--$28\times$, with TimesFM steepest at $27.9$.

\begin{table}[t]
\centering
\caption{Mean faulted error $\mathrm{MSE_{cor}}$ at each difficulty level (averaged over $6$ datasets $\times\,4$ modes), sorted by $d_{02}$. The slope $d_{10}/d_{02}$ summarizes sensitivity to severity and separates three tiers. Foundation rows shaded.}
\label{tab:difflevels}
\footnotesize
\setlength{\tabcolsep}{4pt}
\renewcommand{\arraystretch}{1.05}
\begin{tabular*}{\columnwidth}{@{\extracolsep{\fill}}l rrrrr r@{}}
\toprule
\textbf{Model} & $d_{02}$ & $d_{04}$ & $d_{06}$ & $d_{08}$ & $d_{10}$ & $d_{10}/d_{02}$ \\
\midrule
GRU            & 0.76 & 0.76 & 0.79  & 0.78  & 0.79  & \textbf{1.0} \\
LSTM           & 0.77 & 0.77 & 0.79  & 0.78  & 0.79  & \textbf{1.0} \\
TCN            & 1.29 & 2.93 & 8.16  & 14.1  & 10.5  & 8.1 \\
N-BEATS        & 2.46 & 9.13 & 16.4  & 30.4  & 49.6  & 20.1 \\
Autoformer     & 2.98 & 9.31 & 21.8  & 38.0  & 48.9  & 16.4 \\
FEDformer      & 3.14 & 9.46 & 21.8  & 38.0  & 49.0  & 15.6 \\
NSTransformers & 5.00 & 18.5 & 44.3  & 86.0  & 122.0 & 24.4 \\
TimesNet       & 5.14 & 18.5 & 44.8  & 82.6  & 115.6 & 22.5 \\
DLinear        & 6.18 & 22.7 & 56.2  & 104.1 & 145.0 & 23.5 \\
NLinear        & 6.63 & 24.4 & 61.6  & 111.7 & 156.7 & 23.7 \\
iTransformer   & 7.51 & 28.0 & 68.0  & 129.1 & 183.2 & 24.4 \\
PatchTST       & 7.86 & 29.3 & 72.3  & 133.4 & 190.3 & 24.2 \\
TimeXer        & 8.01 & 30.2 & 73.3  & 136.3 & 192.1 & 24.0 \\
TimeMixer      & 8.04 & 30.5 & 74.5  & 139.9 & 197.6 & 24.6 \\
ARIMA          & 10.6 & 39.5 & 96.7  & 190.1 & 271.8 & 25.7 \\
SeasonalNaive  & 10.8 & 40.1 & 97.3  & 183.4 & 261.0 & 24.1 \\
ETS            & 11.8 & 43.3 & 103.9 & 196.4 & 278.1 & 23.5 \\
Naive          & 12.2 & 44.3 & 106.2 & 201.0 & 283.6 & 23.2 \\
TimesFM & 12.3 & 51.9 & 135.6 & 271.0 & 342.6 & 27.9 \\
Moirai  & 12.5 & 50.7 & 130.5 & 251.6 & 320.1 & 25.7 \\
Chronos & 14.0 & 55.2 & 136.8 & 263.8 & 357.9 & 25.5 \\
\bottomrule
\end{tabular*}
\end{table}

\subsection{RQ6: Do Data Characteristics Modulate Fragility?}
\label{sec:rq6}
Robustness is primarily a property of the architecture, but dimensionality sets the amplitude. Across the six datasets, the model ordering is highly consistent, so fragility is structural rather than dataset-specific. What dimensionality controls is difficulty: on the $321$-channel Electricity dataset, whose dense cross-channel correlation Modes~II/IV disrupt most, ratios reach their extremes (TimesFM $2090$, TimeXer $1112$), whereas the low-dimensional, milder Weather keeps most ratios in the $1$--$55$ range; even on Electricity, GRU and LSTM remain near $1.2$. Sampling granularity (hourly ETTh vs.\ $15$-minute ETTm) barely changes the ordering, recurrent and convolutional models stay robust, and foundation models stay fragile across both. Table~\ref{tab:perdataset} reports the full per-dataset breakdown for all $21$ models (ARIMA omitted on the $321$-channel Electricity). Every dataset preserves the same broad ordering, while the robustness ratio scales with dimensionality: from the $1$--$55$ band on the low-dimensional Weather, to peaks of $2090$ (TimesFM) and $1112$ (TimeXer) on the $321$-channel Electricity, even as GRU and LSTM stay near $1.2$ everywhere. Full per-dataset instances are additionally released with the benchmark artifacts.

\begin{table*}[t]
\centering
\caption{Per-dataset faulted error and robustness ratio for all $21$ models, grouped by category. For each dataset we report the mean faulted error $\mathrm{MSE_{cor}}$ (averaged over $4$ modes $\times\,5$ difficulties) and the robustness ratio $r=\mathrm{MSE_{cor}}/\mathrm{MSE_{clean}}$ ($r\!=\!1$ is perfect robustness). ARIMA is omitted on the $321$-channel Electricity, where per-series fitting is impractical (``--''). Bold marks the best (lowest) $r$ per dataset.}
\label{tab:perdataset}
\setlength{\tabcolsep}{4pt}
\begin{tabular}{@{}l rr rr rr rr rr rr@{}}
\toprule
\textbf{Model}
& \multicolumn{2}{c}{\textbf{ETTh1}}
& \multicolumn{2}{c}{\textbf{ETTh2}}
& \multicolumn{2}{c}{\textbf{ETTm1}}
& \multicolumn{2}{c}{\textbf{ETTm2}}
& \multicolumn{2}{c}{\textbf{Electricity}}
& \multicolumn{2}{c}{\textbf{Weather}} \\
\cmidrule(lr){2-3}\cmidrule(lr){4-5}\cmidrule(lr){6-7}\cmidrule(lr){8-9}\cmidrule(lr){10-11}\cmidrule(lr){12-13}
& MSE & $r$ & MSE & $r$ & MSE & $r$ & MSE & $r$ & MSE & $r$ & MSE & $r$ \\
\midrule
\rowcolor{black!6}\multicolumn{13}{@{}l}{\textbf{Statistical}} \\
Naive          & 184.7 & 154.3 & 153.1 & 96.5  & 122.5 & 134.1 & 110.5 & 121.1 & 171.3 & 85.4  & 34.7 & 42.6 \\
SeasonalNaive  & 168.1 & 230.9 & 138.2 & 98.1  & 113.2 & 114.6 & 102.7 & 102.3 & 156.6 & 412.2 & 32.3 & 35.0 \\
ARIMA          & 185.2 & 185.0 & 153.2 & 105.4 & 123.0 & 136.5 & 111.8 & 131.9 & --    & --    & 35.4 & 44.8 \\
ETS            & 180.2 & 199.3 & 149.4 & 115.3 & 120.2 & 145.3 & 108.1 & 130.8 & 168.4 & 103.1 & 34.0 & 43.7 \\
\midrule
\rowcolor{black!6}\multicolumn{13}{@{}l}{\textbf{Linear / lightweight}} \\
DLinear        & 91.6  & 149.8 & 75.2  & 90.8  & 58.6  & 105.1 & 73.9  & 138.4 & 85.0  & 675.9 & 16.9 & 23.7 \\
NLinear        & 114.5 & 202.5 & 85.0  & 110.8 & 71.1  & 154.7 & 90.5  & 229.8 & 47.0  & 115.6 & 25.3 & 39.0 \\
N-BEATS        & 25.2  & 93.6  & 11.4  & 39.5  & 7.54  & 43.8  & 12.4  & 57.1  & 54.2  & 38.5  & 18.8 & 55.0 \\
\midrule
\rowcolor{black!6}\multicolumn{13}{@{}l}{\textbf{Recurrent / conv.}} \\
LSTM           & 0.92 & \textbf{1.0} & 0.93 & \textbf{1.0} & 0.71 & \textbf{1.1} & 0.66 & \textbf{1.1} & 0.77 & 1.2          & 0.69 & \textbf{1.0} \\
GRU            & 0.92 & 1.1          & 0.94 & 1.1          & 0.72 & 1.2          & 0.66 & 1.2          & 0.74 & \textbf{1.2} & 0.67 & 1.1 \\
TCN            & 11.6 & 11.1         & 11.3 & 10.9         & 2.84 & 3.7          & 3.40 & 4.6          & 13.8 & 15.2         & 1.35 & 1.8 \\
\midrule
\rowcolor{black!6}\multicolumn{13}{@{}l}{\textbf{Decomposition Transf.}} \\
Autoformer     & 29.2 & 28.3 & 28.0 & 25.0 & 21.9 & 26.0 & 22.4 & 33.1 & 35.7 & 168.0 & 7.96 & 7.6 \\
FEDformer      & 29.3 & 27.2 & 27.6 & 23.3 & 22.8 & 21.7 & 21.8 & 29.8 & 36.2 & 25.3  & 7.94 & 8.3 \\
\midrule
\rowcolor{black!6}\multicolumn{13}{@{}l}{\textbf{Attention / SOTA}} \\
PatchTST       & 102.8 & 190.4 & 79.9 & 95.3  & 99.2  & 179.8 & 95.2 & 178.7 & 113.8 & 891.4  & 28.8 & 40.3 \\
iTransformer   & 128.2 & 250.7 & 77.0 & 96.3  & 92.6  & 199.4 & 75.5 & 144.9 & 104.0 & 914.1  & 21.7 & 28.4 \\
TimeXer        & 129.7 & 247.7 & 81.7 & 96.6  & 82.9  & 170.3 & 77.6 & 150.1 & 134.1 & 1112.2 & 22.0 & 30.1 \\
TimeMixer      & 121.7 & 190.3 & 93.3 & 105.7 & 101.8 & 189.5 & 89.5 & 143.6 & 111.9 & 925.6  & 22.3 & 30.3 \\
TimesNet       & 37.9  & 56.2  & 64.3 & 72.4  & 71.3  & 157.7 & 63.3 & 111.7 & 64.1  & 424.5  & 19.2 & 27.5 \\
NSTransformers & 76.8  & 131.1 & 55.5 & 67.4  & 50.7  & 171.9 & 60.0 & 119.0 & 71.2  & 91.6   & 16.8 & 25.3 \\
\midrule
\rowcolor{black!6}\multicolumn{13}{@{}l}{\textbf{Foundation, zero-shot}} \\
TimesFM        & 233.8 & 334.2 & 187.5 & 201.9 & 153.9 & 298.4 & 142.6 & 258.3 & 214.2 & 2090.0 & 44.0 & 148.2 \\
Chronos        & 243.4 & 371.2 & 193.6 & 200.9 & 157.6 & 290.4 & 138.9 & 237.2 & 214.7 & 1920.1 & 45.1 & 55.3 \\
Moirai         & 223.0 & 315.2 & 181.2 & 181.9 & 146.9 & 210.6 & 130.7 & 180.2 & 193.6 & 1252.5 & 43.0 & 52.9 \\
\bottomrule
\end{tabular}%
\end{table*}

\medskip
\noindent Taken together, RQ1--RQ6 show that no model occupies the accurate-and-robust regime: the strongest clean forecasters are the most fragile, the most robust are only average on clean data, and the entire failure budget concentrates in the mechanism-level modes that clean-data leaderboards cannot see.

\section{Discussion}
\label{sec:discussion}

\subsection{Model-Selection Risk}
\label{sec:msr}
Sections~\ref{sec:rq1} and \ref{sec:rq2} together expose a concrete hazard we call model-selection risk. Clean MSE is a trustworthy selection criterion only under observation-level faults, where the clean and faulted rankings agree ($\rho>0.92$); under the mechanism-level faults that any real deployment eventually meets, the two rankings are uncorrelated (Modes~III/IV, $\rho<0.06$) and, taken globally, anti-correlated ($\rho=-0.544$). Choosing a forecaster by its position on a clean leaderboard, the prevailing practice, can therefore be worse than choosing at random in a deployment dominated by regime change or cascading sensor faults, because the clean ranking actively points toward the most fragile architectures (Table~\ref{tab:overall}). The corollary is methodological: a model's value cannot be summarized by a single clean-accuracy number but should be reported as a joint claim over clean-data accuracy and structural robustness, stratified by mode and difficulty. This matters beyond any one deployment, because evaluation metrics steer research~\cite{ribeiro2020beyond,kiela2021dynabench}, a community that rewards only clean held-out error will keep producing architectures optimized for it, leaving the fault robustness that deployment requires unmeasured and unimproved.

\subsection{Guarding Against Goodhart}
\label{sec:goodhart}
A benchmark that exposes its scenario parameters invites an obvious objection~\cite{manheim2018categorizing}:
Once $\Theta$ is public, models will be tuned to the values TS-Fault uses, and reported robustness rises without real improvement. The objection is valid but overlooks a defense unavailable to accuracy-centric benchmarks. Because $\Theta$ is compositional (Sec.~\ref{sec:modes}, Eq.~\eqref{eq:compose}), a release can \emph{hold out} parameter combinations: the parameterization is published while evaluation runs on combinations no participant has seen, and the space of held-out combinations grows with the dimensionality of $\Theta$. This is the time series analogue of the held-out corruptions used to keep ImageNet-C honest~\cite{mintun2021interaction}. Fundamentally, the alternative to exposed parameters is not a benchmark immune to gaming but one whose gaming is invisible, an opaque test set does not prevent overfitting, it only hides it. Transparency about $\Theta$ is a feature, not a vulnerability.

\subsection{Relation to Adjacent Problems}
\label{sec:adjacent}
TS-Fault borders several established problems but answers a different question in each case.
\emph{(i) Anomaly detection}~\cite{chandola2009anomaly,schmidl2022anomaly,xu2021anomaly,audibert2020usad} asks whether a sample is anomalous; TS-Fault asks whether a forecaster still produces a trustworthy forecast when an anomaly is its input. The former treats the anomaly as a label, the latter as an operator acting on the input. \emph{(ii) Change-point detection}~\cite{aminikhanghahi2017survey,keogh2005hot} locates where processes change; Modes~III/IV invert this, we inject a change point at a known location and test whether it misleads the forecaster. \emph{(iii) Concept drift}~\cite{yao2022wild,gagnon2022woods} studies online adaptation as a distribution slowly shifts; TS-Fault is an offline protocol that measures robustness without retraining. The two are complementary. \emph{(iv) Adversarial robustness}~\cite{madry2017towards,goodfellow2014explaining,liu2022practical} seeks the worst-case perturbation inside an $\epsilon$-ball: a tight bound, but a physically meaningless one. TS-Fault replaces the worst-case search with semantically interpretable operators, trading ``how much adversarial noise can a model absorb?'' for ``under which named failure mechanism does it break?'' \emph{(v) Distribution-shift benchmarks}~\cite{koh2021wilds,yao2022wild,gagnon2022woods} treat the train--test gap as a single static object; TS-Fault separates its structural subtypes, regime transition, dependency fracture, cascading propagation, and lets a single evaluation traverse them at controlled difficulty.

\subsection{Threats to Validity}
\label{sec:threats}
\noindent\emph{(i)Ecological validity:} our faults are constructed, not harvested from incident logs; we mitigate this by grounding each mode in documented failures~\cite{busby2021cascading,forbes2002no,moor2021early} and, as in ImageNet-C~\cite{hendrycks2019benchmarking}, by locating validity in structural fidelity rather than historical origin (Sec.~\ref{sec:imagenetc}), leaving large-scale field validation to future work. 
\emph{(ii)Hyperparameters:} the weights $\lambda$ and $\beta$ are fixed per mode, yet degradation is monotone in difficulty (Sec.~\ref{sec:rq5}), and the mechanism-level collapse is so complete ($\rho\approx0.03$) that no reweighting plausibly restores the ranking. 
\emph{(iii)Scope:} the ordering is consistent across our six energy, load, and climate datasets (Sec.~\ref{sec:rq6}), suggesting structural conclusions, though finance, traffic, and clinical domains remain untested. 
\emph{(iv)Training budget:} a shared $10$-epoch recipe keeps comparisons fair, and heavier tuning is unlikely to reverse a stratification that tracks architecture rather than fit.

\section{Conclusion}
\label{sec:conclusion}
Time series forecasting now informs decisions where errors cost money, time, and safety~\cite{smuha2025regulation}, yet its models are still ranked by a single number, the average error on clean held-out data, which can be misleading. 
To provide a more robust way of evaluating TSF methods, we design and introduce TS-Fault, a benchmark that replaces the clean test pair with a structured instance generated by an explicit fault operator and carrying semantic parameters $\Theta$ and a controllable difficulty $\delta$.
We evaluated $21$ forecasting models across $6$ datasets, $4$ fault modes, and $5$ difficulties under a paired clean/corrupt protocol, and reached five findings. \emph{(i)}~Clean-data accuracy and structural robustness are anti-correlated ($\rho=-0.544$, $p=0.011$), and the strongest clean models are among the most fragile. \emph{(ii)}~Basic recurrent and convolutional models, namely LSTM and GRU (ratios $\approx\!1$) and TCN ($7.9$), are the most robust by a wide margin, against ratios from the tens to several hundred for every other family, despite only average clean-data accuracy. \emph{(iii)}~Failure impact is stratified. Observation-level faults preserve the clean ranking ($\rho>0.92$) while mechanism-level faults destroy it ($\rho<0.06$). \emph{(iv)}~All $884$ catastrophic failures fall in the two mechanism-level modes. \emph{(v)}~Pretrained foundation models are the extreme case of strong-but-fragile, top-tier on clean data, but worst of all on robustness. 
Overall, TS-Fault provides a novel perspective in evaluating time series forecasting methods, especially the robustness of the prediction models. It could also lead to new research directions to further promote the development of new TSF methods and enhance their generalization capability in real-world applications.

\bibliographystyle{IEEEtran}
\bibliography{ref}

\end{document}